\documentclass{article}

\PassOptionsToPackage{numbers,compress}{natbib}
\usepackage[preprint]{neurips_2026}

\usepackage[utf8]{inputenc} % allow utf-8 input
\usepackage[T1]{fontenc}    % use 8-bit T1 fonts
\usepackage{hyperref}       % hyperlinks
\usepackage{url}            % simple URL typesetting
\usepackage{booktabs}       % professional-quality tables
\usepackage{amsfonts}       % blackboard math symbols
\usepackage{nicefrac}       % compact symbols for 1/2, etc.
\usepackage{microtype}      % microtypography
\usepackage{xcolor}         % colors

\usepackage{amsmath}
\usepackage{graphicx}
\usepackage{multirow}
\usepackage{subcaption}

\title{PhaseLoRA: Control-Regime-Conditioned Low-Rank Adaptation for Continuous-Action Vision-Language-Action Policies}

\author{%
  Yufei Guo\\
  Tsinghua University\\
  \texttt{guoyufei21@mails.tsinghua.edu.cn} \\
  \And
  Yinan Wu \\
  Tsinghua University \\
  \texttt{yinanwu@tsinghua.edu.cn} \\
  \AND
  Haoran Duan \\
  Tsinghua University \\
  \texttt{haoranduan@tsinghua.edu.cn} \\
  \And
  Guiguang Ding \\
  Tsinghua University \\
  \texttt{dinggg@tsinghua.edu.cn} \\
  \And
  Jungong Han\thanks{Corresponding author.} \\
  Tsinghua University \\
  \texttt{jghan@mail.tsinghua.edu.cn} \\
}

\begin{document}

\maketitle

\begin{abstract}
Parameter-efficient fine-tuning (PEFT) is a natural way to adapt pretrained vision-language-action (VLA) policies, but most adapter designs apply temporally static updates throughout a control rollout, overlooking the phase-dependent nature of continuous-action manipulation. Such policies traverse distinct regimes, including approach, contact transition, grasping, transport, and placement, each requiring different adaptation behaviors. We propose \textbf{PhaseLoRA}, a lightweight LoRA parameterization that conditions adaptation at each action-chunk prediction step using two weakly supervised descriptors: fine-control tendency and event/boundary intensity. PhaseLoRA modulates the LoRA left factor in the action expert, allowing the effective low-rank update direction to vary over time while keeping the backbone largely frozen. On LIBERO, PhaseLoRA improves average success rate by 12.2 points over a matched-parameter high-rank LoRA baseline and outperforms stronger LoRA variants. Ablations show that random temporal modulation and scalar gating do not reproduce the performance of the full model, while update-direction analyses reveal structured temporal variation associated with the predicted control descriptors. These results establish within-trajectory conditioning as an effective lightweight PEFT axis for continuous-action VLA policies.
\end{abstract}

\noindent\textbf{Code:} The code will be released at
\url{https://github.com/Grinffin/PhaseLoRA}.

\section{Introduction}

Pretrained vision-language-action (VLA) policies unify visual perception, language conditioning, and action generation, making them strong starting points for robot manipulation~\citep{BrohanBCCDFGHHH23,pmlr-v202-driess23a,pmlr-v229-zitkovich23a,openxembodiment2024,ghosh2024octo,pmlr-v270-kim25c,pmlr-v305-black25a}. As these backbones grow in scale, downstream adaptation increasingly relies on parameter-efficient fine-tuning (PEFT), especially low-rank adaptation (LoRA)~\citep{hu2022lowrank}, to avoid the cost of full fine-tuning~\citep{pmlr-v270-kim25c,Kim2025FineTuningVM}. However, most PEFT methods treat adaptation as static within an execution: once an adapter is chosen for a task, instruction, or example~\citep{ivison-etal-2023-hint,lv-etal-2024-hyperlora,jin2024conditionalloraparametergeneration}, the same update is applied throughout the control rollout.

This temporal invariance overlooks a basic property of continuous-action manipulation: a single rollout is not a homogeneous control problem. It may progress through coarse approach, contact or release transitions, grasp execution, transport, and precise placement, as also reflected in skill-structured and contact-rich manipulation studies~\citep{vonhartz2024artimitationlearninglonghorizon,cheng2026tacumimultimodaluniversalmanipulation,doi:10.55092/rl20260001,yu2025forcevlaenhancingvlamodels,zhao2025touch}. These regimes can require different corrections to the pretrained policy even within the same task and instruction. We therefore view within-trajectory control heterogeneity as a missing adaptation axis for PEFT in continuous-action VLA policies: the adapter should remain lightweight, but its effective update should be allowed to change with the evolving control regime.

\begin{figure}
  \centering
  \includegraphics[width=0.8\linewidth]{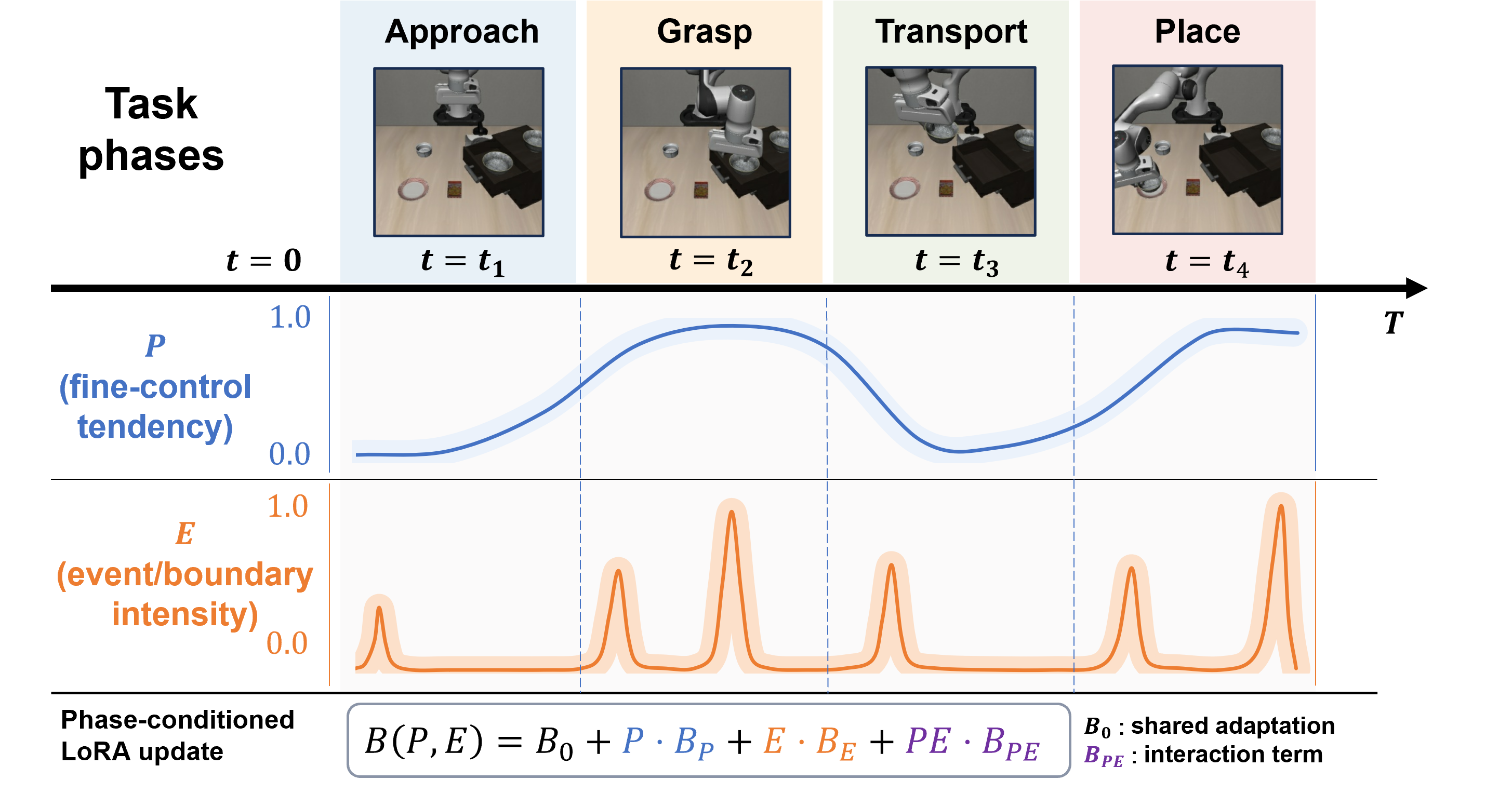}
  \caption{\textbf{Motivation for within-trajectory control-regime conditioning.}
  A single manipulation rollout can traverse different control regimes, including approach, grasp, transport, and precise placement. We summarize these variations with two continuous descriptors: fine-control tendency $P$ and event/boundary intensity $E$. PhaseLoRA uses these descriptors to modulate the LoRA left factor,
  $B(P,E)=B_0 + P\!\cdot\!B_P + E\!\cdot\!B_E + PE\!\cdot\!B_{PE}$,
  so the effective low-rank update can vary across control steps instead of remaining static throughout the rollout.}
  \label{fig:intro_phase}
\end{figure}

We introduce \textbf{PhaseLoRA}, a parameter-efficient adapter that conditions LoRA updates at each action-chunk prediction step using two weakly supervised descriptors: \emph{fine-control tendency} and \emph{event/boundary intensity}. As illustrated schematically in Figure~\ref{fig:intro_phase}, these descriptors are designed to represent precision-demanding moments, such as grasping and placement, and abrupt control changes, such as contact or release. Unlike conditional LoRA methods that adapt across tasks, examples, experts, or instructions~\citep{ivison-etal-2023-hint,lv-etal-2024-hyperlora,jin2024conditionalloraparametergeneration,pmlr-v267-ge25d}, PhaseLoRA conditions adaptation within a single rollout. It modulates the LoRA left factor with shared, descriptor-specific, and interaction components, letting the effective low-rank update direction vary over time while keeping the backbone largely frozen. The descriptors are learned from trajectory statistics, without manual phase labels or full-capacity expert branches.

We evaluate whether within-trajectory conditioning provides benefits beyond standard explanations such as increased parameter count or generic temporal variation. On LIBERO~\citep{liu2024libero}, PhaseLoRA consistently improves over standard LoRA, a matched-parameter high-rank LoRA baseline, and stronger LoRA variants. Through ablations and update-direction analyses, we test whether control-informed descriptors outperform random temporal modulation, whether scalar rescaling of a fixed LoRA direction is sufficient, and whether the learned low-rank update directions vary near salient manipulation events. We further evaluate PhaseLoRA on real-world manipulation tasks, providing supporting evidence that the observed benefits extend beyond simulation to physical execution.

Our contributions are:
\begin{itemize}
    \item We formulate within-trajectory control heterogeneity as a missing adaptation axis for PEFT in continuous-action VLA policies.
    \item We introduce \textbf{PhaseLoRA}, a lightweight LoRA parameterization that conditions the adapter on weakly supervised fine-control and event/boundary descriptors while preserving low-rank updates at each action-chunk prediction step.
    \item We show that PhaseLoRA outperforms the matched-parameter high-rank LoRA, random-modulation, and scalar-gating controls, and analyze how its effective low-rank update directions vary with the predicted descriptors and near salient manipulation events.
\end{itemize}

\section{Related Work}

\paragraph{Generalist robot policies and vision-language-action models.}
Generalist robot learning has advanced from multitask imitation policies such as BC-Z~\cite{pmlr-v164-jang22a} and PerAct~\cite{pmlr-v205-shridhar23a} to large-scale robot foundation models trained with multimodal and multi-embodiment data. RT-1~\cite{BrohanBCCDFGHHH23} showed that scaling data and model capacity can improve real-world robotic control, while PaLM-E~\cite{pmlr-v202-driess23a} and RT-2~\cite{pmlr-v229-zitkovich23a} connected pretrained language or vision-language models to embodied action generation. Recent systems such as RT-X~\cite{openxembodiment2024}, Octo~\cite{ghosh2024octo}, OpenVLA~\cite{pmlr-v270-kim25c}, $\pi_{0.5}$~\cite{pmlr-v305-black25a}, and RDT-1B~\cite{Liu2024RDT1BAD} further establish pretrained VLA backbones as practical starting points for robot manipulation. Our work builds on this setting and focuses on how to adapt such continuous-action VLA policies efficiently to downstream manipulation tasks.

\paragraph{Fine-tuning and parameter-efficient adaptation for robot foundation models.}
As robot foundation models scale, efficient downstream adaptation has become a practical bottleneck. OpenVLA~\cite{pmlr-v270-kim25c} identified efficient fine-tuning as important for VLA deployment, and OpenVLA-OFT~\cite{Kim2025FineTuningVM} systematically studied adaptation choices such as decoding, action representation, action chunking, and training objectives. Parameter-efficient VLA adaptation has also been explored through LoRA-style tuning, input- and layer-wise adaptive rank allocation~\cite{kim2026adaptive}, and sequential LoRA adaptation in continual RL~\cite{hu2026simplerecipeworksvisionlanguageaction}. RL-based post-training methods such as CO-RFT~\cite{huang2025corft} and VLA-RFT~\cite{li2025vlarftvisionlanguageactionreinforcementfinetuning} are complementary to our setting, as they mainly modify the learning signal rather than the adapter parameterization. Our focus is also complementary: rather than adapting rank or the post-training objective, we condition the LoRA update direction on weakly supervised control descriptors within a single rollout for continuous-action VLA fine-tuning.

\paragraph{LoRA variants, dynamic adapters, and conditional parameter generation.}
Beyond robotics, LoRA has been extended through rank or budget allocation, such as DyLoRA~\cite{valipour-etal-2023-dylora} and AdaLoRA~\cite{zhang2023adalora}, alternative parameterizations such as DoRA~\cite{pmlr-v235-liu24bn}, and mixtures of LoRA experts such as X-LoRA~\cite{buehler2024xlora_journal} and D-MoLE~\cite{pmlr-v267-ge25d}. Structured PEFT methods further show that a single shared low-rank update can be limiting under heterogeneous adaptation demands: MokA~\cite{wei2025moka} separates unimodal and cross-modal adaptation, while HINT~\cite{ivison-etal-2023-hint}, HyperLoRA~\cite{lv-etal-2024-hyperlora}, and Conditional LoRA Parameter Generation~\cite{jin2024conditionalloraparametergeneration} generate adapter parameters conditionally. Our distinction is within-rollout conditioning at each policy query for continuous-action VLA policies. Prior conditional LoRA methods mainly adapt across tasks, modalities, instructions, experts, or input examples; PhaseLoRA instead modulates the action-expert LoRA update once per action-chunk prediction within a single rollout, using weakly supervised descriptors of the current control regime. This targets temporal control heterogeneity in manipulation without requiring explicit phase labels or full-capacity expert branches.

\paragraph{Phase structure, skill decomposition, and contact-rich manipulation.}
Another line of work observes that manipulation is rarely a homogeneous control process. Long-horizon imitation methods often decompose demonstrations into skills or introduce intermediate representations, as in The Art of Imitation~\cite{vonhartz2024artimitationlearninglonghorizon}, TacUMI~\cite{cheng2026tacumimultimodaluniversalmanipulation}, and RoboInter~\cite{li2026robointerholisticintermediaterepresentation}. Contact-rich manipulation methods model interaction regimes more explicitly, for example by separating reaching from local interaction~\cite{zhao2025touch} or conditioning policies on force or tactile information, as in TLA~\cite{doi:10.55092/rl20260001}, ForceVLA~\cite{yu2025forcevlaenhancingvlamodels}, Tactile-VLA~\cite{huang2025tactilevlaunlockingvisionlanguageactionmodels}, and HapticVLA~\cite{gubernatorov2026hapticvlacontactrichmanipulationvisionlanguageaction}. Concurrent VLA methods have also introduced phase-conditioned action generation: BehaviorVLA uses a phase-conditioned behavior decoder, while Mag-VLA combines a motion-aware phase classifier with a phase-conditioned Action Chunking Transformer decoder~\cite{hu2026behaviorvla,wang2026magvla}. These works motivate our view that a single rollout can contain heterogeneous control regimes. We differ by conditioning the low-rank adapter rather than the action decoder, without requiring explicit skill boundaries, manual phase labels, or force or tactile measurements.

\section{Method}
\label{sec:method}

\subsection{Setup}
\label{subsec:setup}

We study parameter-efficient fine-tuning for continuous-action vision-language-action (VLA) policies. Let $t$ denote a low-level control timestep at which the policy is queried, and let $x_t$ denote the policy context available at that query, including the observation, language instruction, and any additional state inputs used by the backbone. A pretrained VLA policy predicts an action chunk
\begin{equation}
    \mathbf{a}_{t:t+H-1} = f_{\theta}(x_t),
\end{equation}
where $H$ is the action horizon and $\theta$ denotes the pretrained policy parameters.

We consider low-rank adaptation while keeping the pretrained backbone frozen except for a small set of trainable adaptation parameters. For a target linear layer with pretrained weight $W \in \mathbb{R}^{d_{\mathrm{out}} \times d_{\mathrm{in}}}$, standard LoRA replaces $W$ with
\begin{equation}
    W' = W + \Delta W, 
    \qquad
    \Delta W = BA,
\end{equation}
where $B \in \mathbb{R}^{d_{\mathrm{out}} \times r}$ and $A \in \mathbb{R}^{r \times d_{\mathrm{in}}}$ are trainable low-rank factors, and $r$ is the adaptation rank. For clarity, we omit the standard LoRA scaling factor in notation.

Standard LoRA applies the same update matrix $\Delta W$ at every policy query. This is restrictive for continuous-action manipulation, where a single rollout can traverse substantially different control regimes, from coarse motion and contact transitions to grasp execution and precise alignment. Our goal is therefore to replace the static update with a within-trajectory, control-regime-conditioned low-rank adaptation while retaining the efficiency of LoRA.

\begin{figure}
  \centering
  \includegraphics[width=1.0\linewidth]{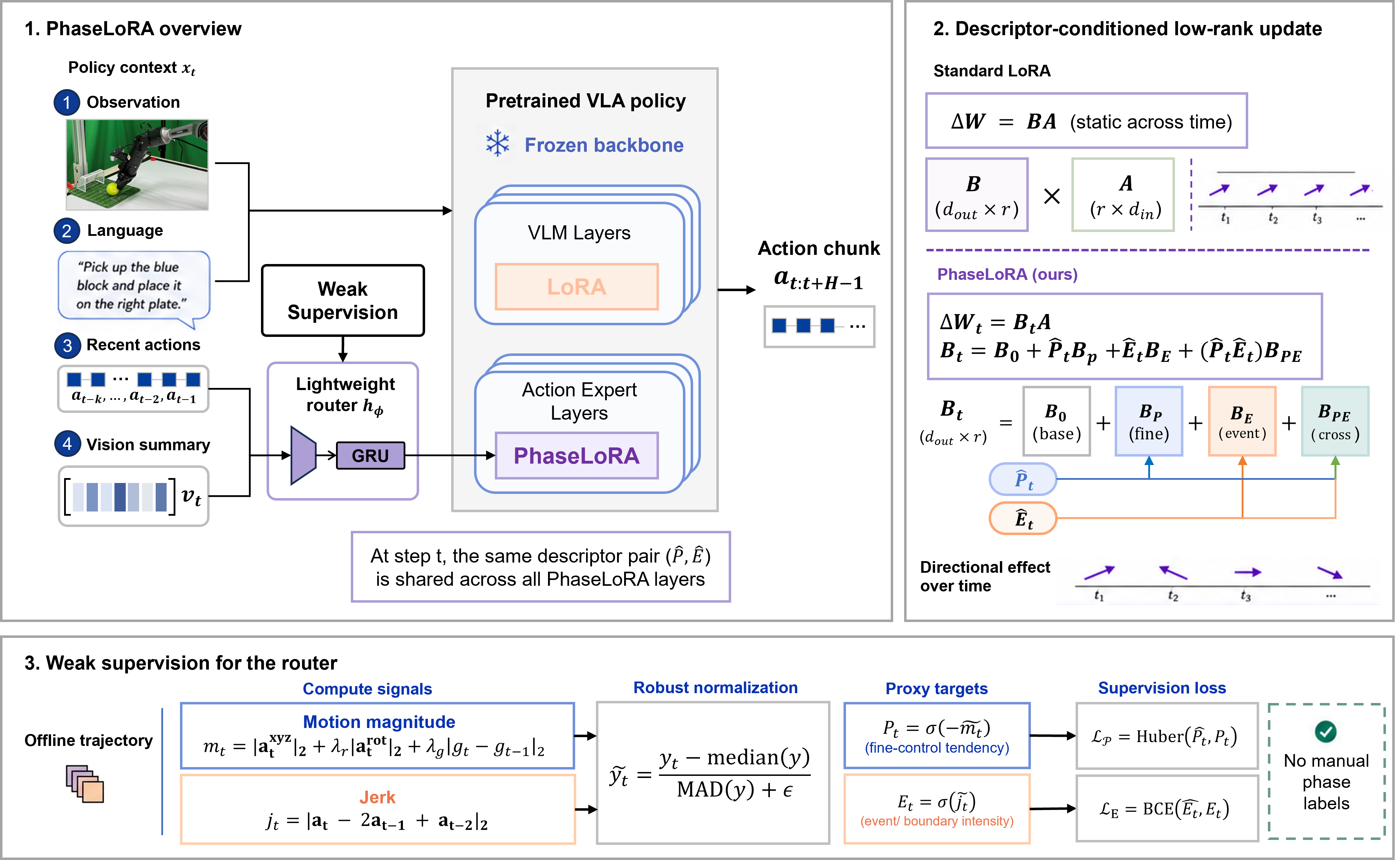}
  \caption{\textbf{Overview of PhaseLoRA.} A lightweight router predicts fine-control tendency $\hat P_t$ and event/boundary intensity $\hat E_t$ from the current policy context. These descriptors modulate the action-expert LoRA update through shared, descriptor-specific, and interaction components, yielding control-regime-dependent low-rank corrections within a rollout. The router is trained with weak proxy targets derived from trajectory statistics.}
  \label{fig:method}
\end{figure}

\subsection{PhaseLoRA}
\label{subsec:phase_lora}

Our key idea is to make the LoRA update depend on the current control regime within a trajectory. At a policy query issued at timestep $t$, a lightweight router predicts two scalar control descriptors $(\hat P_t,\hat E_t) \in [0,1]^2$ for the forward pass that produces the action chunk $\mathbf{a}_{t:t+H-1}$. The same descriptor pair is shared across all PhaseLoRA layers for that forward pass, while each layer keeps its own low-rank parameters.

We apply descriptor-conditioned adaptation only to the policy's action-prediction module, referred to as the \emph{action expert}; any VLM-side LoRA parameters remain standard and time-invariant. For an adapted action-expert layer with pretrained weight $W \in \mathbb{R}^{d_{\mathrm{out}} \times d_{\mathrm{in}}}$, PhaseLoRA replaces the static LoRA update with
\begin{equation}
    \Delta W_t = B_t A,
    \label{eq:delta_main}
\end{equation}
where $A \in \mathbb{R}^{r \times d_{\mathrm{in}}}$ is the trainable right factor for that layer and is shared across policy queries. The left factor is conditioned on the predicted descriptors:
\begin{equation}
    B_t
    =
    B_0
    + \hat P_t B_P
    + \hat E_t B_E
    + (\hat P_t \hat E_t) B_{PE},
    \label{eq:B_full}
\end{equation}
with $B_0, B_P, B_E, B_{PE} \in \mathbb{R}^{d_{\mathrm{out}} \times r}$.

The shared component $B_0$ parameterizes the time-invariant part of the adapter, while $B_P$, $B_E$, and $B_{PE}$ parameterize descriptor-dependent deviations associated with fine-control tendency, event/boundary intensity, and their interaction. We condition only the left factor: the shared right factor $A$ defines a common input projection, whereas the descriptor-conditioned $B_t$ changes the output directions of the low-rank correction. For any fixed $(\hat P_t,\hat E_t)$, the update remains rank at most $r$, since $\Delta W_t$ is still the product of a $d_{\mathrm{out}}\times r$ matrix and an $r\times d_{\mathrm{in}}$ matrix. Thus, PhaseLoRA preserves the low-rank structure of LoRA while allowing the effective update direction to vary across policy queries, rather than only rescaling a fixed update.

\subsection{Weak Supervision for Control Descriptors}
\label{subsec:weak_phase}

The control descriptors are not directly observed. Instead, during training, we construct weak proxy targets for two interpretable quantities: a fine-control tendency score $P_t \in [0,1]$ and an event/boundary intensity score $E_t \in [0,1]$. These targets are computed offline from the low-level action and gripper-command sequence of each training trajectory, and used only to supervise the router.

\paragraph{Fine-control tendency.}
We define a motion statistic
\begin{equation}
    m_t
    =
    \|\mathbf{a}^{xyz}_t\|_2
    + \lambda_r \|\mathbf{a}^{rot}_t\|_2
    + \lambda_g |g_t - g_{t-1}|,
    \label{eq:m_def}
\end{equation}
where $\mathbf{a}^{xyz}_t$ and $\mathbf{a}^{rot}_t$ are the translational and rotational components of the low-level action at step $t$, $g_t$ is the gripper command, and $\lambda_r,\lambda_g$ are fixed coefficients used to balance terms with different scales. Lower motion magnitude is treated as a weak indicator of finer-grained control.

\paragraph{Event/boundary intensity.}
To capture abrupt control changes, we define
\begin{equation}
    j_t = \|\mathbf{a}_t - 2\mathbf{a}_{t-1} + \mathbf{a}_{t-2}\|_2,
    \label{eq:ju_def}
\end{equation}
where $j_t$ measures action jerk.

For each trajectory, we robustly normalize each statistic over time. For $y \in \{m,j\}$, let $y$ also denote the corresponding per-trajectory time series; we define
\begin{equation}
    \tilde y_t
    =
    \frac{y_t - \mathrm{median}(y)}
         {\mathrm{MAD}(y) + \epsilon},
    \label{eq:robust_norm}
\end{equation}
where $\mathrm{MAD}(\cdot)$ denotes the median absolute deviation. We then form the weak proxy targets as
\begin{equation}
    P_t = \sigma(-\tilde m_t),
    \qquad
    E_t = \sigma(\tilde j_t),
    \label{eq:pe_targets}
\end{equation}
where $\sigma(\cdot)$ is the sigmoid function.

We emphasize that $P_t$ and $E_t$ are weak proxy targets, not exact symbolic phase labels. Their purpose is not to recover a unique segmentation of the task, but to provide low-dimensional supervision for learning continuous control descriptors that approximately track fine-control tendency and event intensity.

\subsection{Router and Training}
\label{subsec:router_training}

Since the weak proxy targets $(P_t,E_t)$ are unavailable at inference time, we train a lightweight router $h_\phi$ to predict descriptors from information available up to control step $t$:
\begin{equation}
    (\hat P_t, \hat E_t) = h_\phi(\mathbf{z}_t),
    \qquad
    \hat P_t,\hat E_t \in [0,1].
\end{equation}
The router input $\mathbf{z}_t$ includes a visual-semantic summary from the VLA backbone, recent action and gripper history. The history branch uses the previous $K=6$ control steps and retains the first $5$ executed low-level actions from each past action chunk. During training, the history is taken from offline past actions; during inference, it is updated online from previously executed model outputs. The router is a one-layer GRU with hidden dimension $128$ followed by two sigmoid heads; further details are in the appendix.

We jointly optimize PhaseLoRA and the router. Let $\mathcal{L}_{\mathrm{policy}}$ be the original VLA action-prediction loss. The router is supervised by the weak targets from Sec.~\ref{subsec:weak_phase}:
\begin{equation}
    \mathcal{L}_P = \mathrm{Huber}(\hat P_t, P_t),
    \qquad
    \mathcal{L}_E = \mathrm{BCE}(\hat E_t, E_t),
\end{equation}
with losses averaged over valid control steps. The total objective is
\begin{equation}
    \mathcal{L}
    =
    \mathcal{L}_{\mathrm{policy}}
    + \lambda_P \mathcal{L}_P
    + \lambda_E \mathcal{L}_E
    + \mathcal{L}_{\mathrm{reg}},
    \label{eq:total_loss}
\end{equation}
where $\lambda_P$ and $\lambda_E$ weight the auxiliary supervision terms. We regularize descriptor-dependent LoRA components across PhaseLoRA layers:
\begin{equation}
    \mathcal{L}_{\mathrm{reg}}
    =
    \sum_{\ell \in \mathcal{S}}
    \left(
    \lambda_P^{\mathrm{reg}} \|B_{P,\ell}\|_F^2
    +
    \lambda_E^{\mathrm{reg}} \|B_{E,\ell}\|_F^2
    +
    \lambda_{PE}^{\mathrm{reg}} \|B_{PE,\ell}\|_F^2
    \right),
\end{equation}
where $\mathcal{S}$ is the set of PhaseLoRA layers. We initialize $B_P$, $B_E$, and $B_{PE}$ to zero and initialize $(B_0,A)$ with the standard LoRA initialization, so each layer starts as ordinary LoRA and learns descriptor-dependent deviations during fine-tuning.

\subsection{Complexity}
\label{subsec:complexity}

Compared with standard LoRA, our method introduces additional left factors while keeping the right factor fixed across control steps. For a layer with input dimension $d_{\mathrm{in}}$, output dimension $d_{\mathrm{out}}$, and rank $r$, standard LoRA uses $r(d_{\mathrm{in}} + d_{\mathrm{out}})$ trainable parameters. Our method uses $r d_{\mathrm{in}} + 4 r d_{\mathrm{out}}$, corresponding to one right factor $A$ and four left factors $(B_0, B_P, B_E, B_{PE})$. Relative to standard LoRA, the additional parameter cost is therefore $3 r d_{\mathrm{out}}$ per adapted layer, plus the lightweight router.

At inference time, the extra computation comes from (i) evaluating the router and (ii) computing the descriptor-conditioned left factor in Eq.~\eqref{eq:B_full}. No additional full-capacity expert branches are introduced. Thus, relative to standard LoRA, our method adds only a small descriptor-dependent overhead, while remaining substantially more efficient than full fine-tuning.

\section{Experiments}
\label{sec:experiments}

We evaluate PhaseLoRA with four goals: whether within-trajectory conditioning improves over static LoRA, whether the gain can be explained by increased adaptation capacity, whether descriptor-conditioned updates are necessary, and whether the learned descriptors align with their intended control roles.

\subsection{Experimental Setup}
\label{subsec:exp_setup}

\paragraph{Benchmarks.}
Our primary controlled benchmark is LIBERO~\citep{liu2024libero}, where we report the main comparisons, ablations, and descriptor analyses. We evaluate on the standard LIBERO suites and report both suite-level success rates and their macro average. Unless otherwise stated, ablations and descriptor analyses are conducted on LIBERO.

\paragraph{Real-world benchmark.}
We additionally evaluate on a four-task real-world tabletop manipulation benchmark collected with a Piper robotic arm and trained using the LeRobot~\citep{cadene2026lerobot} framework. The tasks require moving a tennis ball from a yellow plate to a blue plate, removing a cuboid from a blue plate, picking a red cube into a yellow plate, and putting an apple on a yellow plate. Each task provides 100 demonstrations for fine-tuning, and each method is evaluated with 10 trials per task for each of 3 random seeds, resulting in 30 trials per task per method. We report the mean and standard deviation across the 3 seeds.

\paragraph{Backbones and adaptation setting.}
Unless otherwise stated, our main experiments use the pretrained $\pi_{0.5}$  backbone. Within each backbone, all methods use the same observation and language inputs, action representation, action chunk length, training data, and adapted module family. To isolate the effect of the adaptation mechanism, each PEFT method is applied to the same subset of layers. For PhaseLoRA, descriptor-conditioned adaptation is applied only to the action expert, following Sec.~\ref{subsec:phase_lora}; VLM-side LoRA parameters, when present, remain standard and time-invariant.

\paragraph{Compared methods.}
We compare PhaseLoRA against five PEFT baselines: \textbf{LoRA}~\citep{hu2022lowrank}, standard low-rank adaptation with a static update; \textbf{High-rank LoRA}, a larger-rank baseline chosen to approximately match PhaseLoRA's trainable-parameter budget; \textbf{DoRA}~\citep{pmlr-v235-liu24bn}, a weight-decomposed low-rank adaptation baseline; \textbf{LoRA-MoE}~\cite{pmlr-v267-ge25d}, a mixture-based low-rank adaptation baseline; and \textbf{LoRA-SP}~\citep{kim2026adaptive}, a structured LoRA baseline for VLA fine-tuning. High-rank LoRA controls for trainable capacity, while DoRA, LoRA-MoE, and LoRA-SP test whether stronger static or structured adapter variants recover the same gains.

\paragraph{Training and metrics.}
For each benchmark and backbone, all methods are trained with the same optimizer, batch size, data mixture, and number of training steps. Unless otherwise stated, LIBERO results are averaged over three random seeds, and real-world results are averaged over repeated trials for each task. We report episode-level success rate and the percentage of trainable parameters for each PEFT method. Additional implementation details are provided in the appendix.

\subsection{Main Results on LIBERO}
\label{subsec:libero_main}

Table~\ref{tab:libero_main} reports the main LIBERO comparison. PhaseLoRA achieves the highest macro-average success rate, improving over standard LoRA by 30.6 points and over matched-parameter High-rank LoRA by 12.2 points. The comparison to standard LoRA tests whether a static low-rank update is sufficient, while the comparison to High-rank LoRA controls for trainable adaptation capacity. PhaseLoRA also outperforms DoRA, LoRA-MoE, and LoRA-SP, suggesting that stronger static or structured LoRA variants do not fully recover the benefit of within-trajectory control-regime conditioning.

\begin{table*}[t]
    \centering
    \caption{
    Main results on LIBERO. We report success rates (\%) on the four standard LIBERO suites and their overall average, together with the number of trainable parameters for each PEFT method. Results are reported as mean $\pm$ standard deviation over 3 random seeds.
    }
    \label{tab:libero_main}
    \small
    \begin{tabular}{l c cccc c}
        \toprule
        \multirow{2}{*}{Method} & \multirow{2}{*}{\#Params} & \multicolumn{5}{c}{LIBERO(\%)} \\
        \cmidrule(lr){3-7}
        & & Spatial & Object & Goal & 10 & Avg. \\
        \midrule
        LoRA~\citep{hu2022lowrank}
        & 2.70\% & $60.7 \pm 0.6$ & $41.3 \pm 2.1$ & $32.7 \pm 1.5$ & $18.7 \pm 0.6$ & $38.3 \pm 1.2$ \\

        High-rank LoRA & 4.42\% & $71.0 \pm 1.0$ & $73.3 \pm 1.2$ & $55.3 \pm 3.1$ & $27.0 \pm 4.6$ & $56.7 \pm 2.3$ \\

        DoRA~\citep{pmlr-v235-liu24bn} & 4.42\% & $72.7 \pm 0.6$ & $74.0 \pm 2.6$ & $57.3 \pm 2.5$ & $29.7 \pm 3.5$ & $58.4 \pm 2.2$ \\

        LoRA-MoE~\citep{pmlr-v267-ge25d} & 4.47\% & $73.3 \pm 1.5$ & $70.7 \pm 2.5$ & $58.0 \pm 2.0$ & $31.0 \pm 4.6$ & $58.3 \pm 2.4$ \\

        LoRA-SP~\citep{kim2026adaptive} & 4.42\% & $76.3 \pm 3.2$ & $72.0 \pm 2.0$ & $56.0 \pm 1.0$ & $32.3 \pm 2.5$ & $59.2 \pm 2.3$ \\

        \midrule
        Ours & 4.41\% & $\mathbf{85.3 \pm 1.2}$ & $\mathbf{88.0 \pm 2.0}$ & $\mathbf{64.0 \pm 2.0}$ & $\mathbf{38.3 \pm 3.2}$ & $\mathbf{68.9 \pm 2.4}$ \\
        \bottomrule
    \end{tabular}
\end{table*}

\subsection{Real-World Evaluation}
\label{subsec:real_main}

We further evaluate PhaseLoRA on a real-world manipulation benchmark collected on our robot platform. Because physical evaluation is costly, we compare against the most relevant baseline: matched-parameter High-rank LoRA. This setting tests whether the trend observed on LIBERO also appears under physical execution, and whether any improvement is explained solely by increased adaptation capacity.

Table~\ref{tab:real_main} reports per-task and average success rates. PhaseLoRA achieves the highest average success rate, improving over High-rank LoRA by 18.4 points. These results provide supporting evidence that within-trajectory descriptor-conditioned adaptation can improve continuous-action VLA fine-tuning beyond the controlled simulation benchmark.

\begin{table}[t]
    \centering
    \small
    \caption{Main results on the real-world benchmark. We report mean $\pm$ standard deviation across 3 seeds, with 10 trials per task for each seed.}
    \label{tab:real_main}
    \begin{tabular}{l cccccc}
        \toprule
        Method & \#Params & Move & Remove & Pick & Put & Avg. \\
        \midrule
        High-rank LoRA & 4.42\% & $60.0 \pm 8.2$ & $53.3 \pm 4.8$ & $63.3 \pm 9.4$ & $26.7 \pm 4.7$ & $50.8 \pm 4.2$\\
        Ours & 4.41\% & $\mathbf{66.7 \pm 4.8}$ & $\mathbf{70.0 \pm 8.2}$ & $\mathbf{90.0 \pm 8.2}$ & $\mathbf{50.0 \pm 8.2}$ & $\mathbf{69.2 \pm 3.1}$\\
        \bottomrule
    \end{tabular}
\end{table}

\subsection{Ablations on LIBERO}
\label{subsec:ablations}

The main LIBERO comparison shows that PhaseLoRA improves over static LoRA, including a matched-parameter High-rank LoRA baseline. We further ablate the design choices of PhaseLoRA on LIBERO-Spatial. These ablations test whether the improvement requires meaningful descriptor alignment, whether weak proxy supervision is useful, whether the two descriptors and their interaction contribute, and whether descriptor-dependent scalar rescaling is sufficient.

\paragraph{Descriptor randomization.}
We first test whether the descriptors must be aligned with the underlying control regime. In the random-descriptor control, the descriptor values used by the PhaseLoRA layers are replaced with pseudo-random values sampled from $\mathrm{Uniform}(0,1)$ for both $P$ and $E$. This preserves time-varying scalar inputs and the PhaseLoRA parameterization, but removes the relationship between the descriptors and the current rollout. If the full model outperforms this control, the gain cannot be explained merely by additional parameters or arbitrary temporal modulation.

\paragraph{Scalar-gating control.}
We next test whether descriptor-dependent rescaling of a fixed LoRA update is sufficient. This baseline uses the same router and predicted descriptors $(\hat P_t,\hat E_t)$ as PhaseLoRA, but constrains the update to be a scalar multiple of a fixed low-rank direction:
\begin{equation}
    \Delta W_t = g_t BA, \qquad
    g_t = \exp(\alpha_P \hat P_t + \alpha_E \hat E_t + \alpha_{PE}\hat P_t\hat E_t),
    \label{eq:scalar_gate}
\end{equation}
where $A$ and $B$ are standard LoRA factors, and $\alpha_P,\alpha_E,\alpha_{PE}$ are learned scalar coefficients for each adapted layer. The scalar coefficients are initialized to zero, so the baseline starts from ordinary LoRA with $g_t=1$.

This control preserves descriptor-dependent temporal modulation, but prevents the descriptors from changing the low-rank update direction. In contrast, PhaseLoRA uses
$B_t = B_0 + \hat P_t B_P + \hat E_t B_E + \hat P_t \hat E_t B_{PE}$,
which changes the effective left factor. A gap between the scalar-gated baseline and PhaseLoRA therefore tests whether direction-changing adaptation is needed beyond scalar modulation.

\begin{table}[t]
\centering
\caption{
Ablations on LIBERO-Spatial. We report suite-level success rate. Random descriptors test whether arbitrary temporal modulation is sufficient, while scalar-gated updates test whether descriptor-dependent rescaling of a fixed LoRA direction recovers the gain.
}
\label{tab:ablations}
\begin{tabular}{lcc}
\toprule
Method & Success (\%) & Drop vs. Ours \\
\midrule
High-rank LoRA & $71.0 \pm 1.0$ & 14.3 \\
Ours w/ random descriptors & $66.3 \pm 3.2$ & 19.0 \\
Ours w/ scalar-gated update & $73.7 \pm 1.5$ & 11.6 \\
Ours w/o proxy supervision & $78.7 \pm 3.1$ & 6.6 \\
Ours ($P$ only) & $80.7 \pm 1.5$ & 4.6 \\
Ours ($E$ only) & $81.0 \pm 2.0$ & 4.3 \\
Ours w/o $PE$ & $82.3 \pm 0.6$ & 3.0 \\
Ours & $\mathbf{85.3 \pm 1.2}$ & 0.0 \\
\bottomrule
\end{tabular}
\end{table}

\paragraph{Results.}
Table~\ref{tab:ablations} summarizes the ablations. Random descriptors substantially underperform PhaseLoRA, indicating that arbitrary time-varying inputs are not sufficient; the descriptors must remain aligned with rollout-dependent control variation. The scalar-gated variant also underperforms PhaseLoRA, showing that descriptor-dependent rescaling of a fixed LoRA direction does not recover the benefit of changing the effective update direction.

Removing proxy supervision reduces performance, suggesting that the weak targets provide a useful training signal for the router. Using only $P$ or only $E$ also underperforms the full model, indicating that fine-control tendency and event/boundary intensity provide complementary information. Removing the interaction component $B_{PE}$ yields a smaller drop, suggesting that most of the gain comes from additive descriptor conditioning, with the interaction term providing an additional refinement.

Overall, these ablations support the main design of PhaseLoRA: the improvement is not explained by matched parameter count, arbitrary temporal modulation, or scalar rescaling alone. Instead, performance depends on descriptor-conditioned changes to the effective low-rank update.

\subsection{Analysis of PhaseLoRA Updates}
\label{subsec:descriptor_analysis}

\paragraph{Update-direction variation.}
The scalar-gating control in Sec.~\ref{subsec:ablations} tests whether descriptor-dependent rescaling is sufficient at the level of task performance. Analytically, scalar-gated LoRA cannot change the update direction: if $\Delta W_t = g_t BA$ and $g_t > 0$, then adjacent updates remain in the same Frobenius direction. We therefore analyze whether PhaseLoRA produces structured update-direction changes over time.

\begin{figure*}[t]
    \centering
    \includegraphics[width=1.0\textwidth]{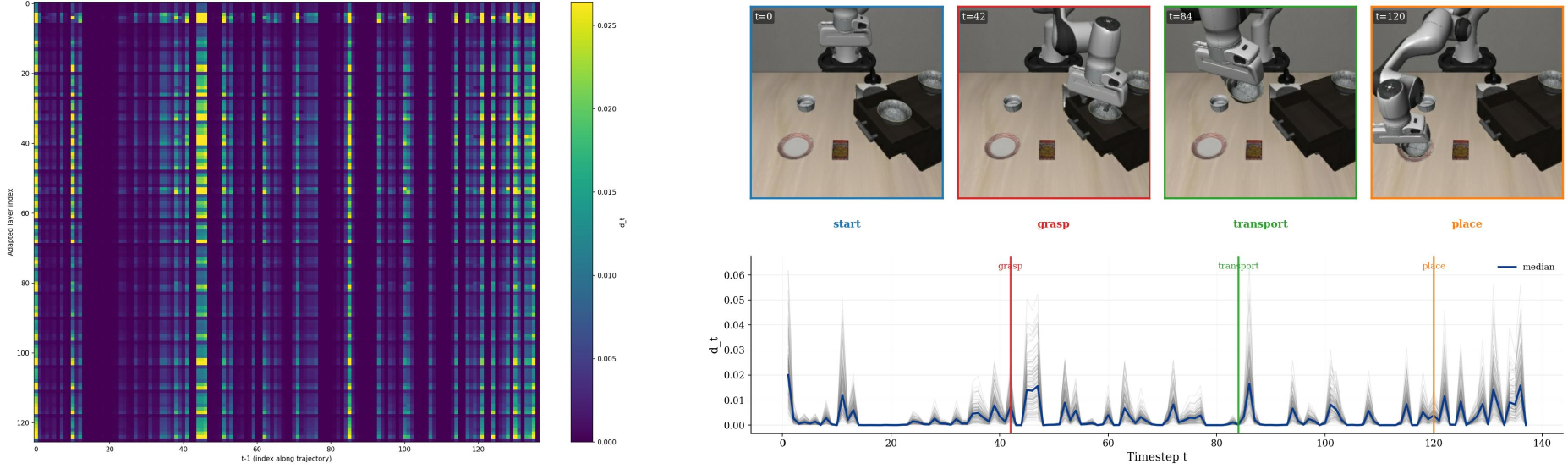}
    \caption{\textbf{Update-direction variation in PhaseLoRA on LIBERO-Spatial.} For one representative rollout, the left panel shows the adjacent-step update-direction distance $d_t$ across adapted layers and timesteps. The upper-right row shows four rollout frames, and the lower-right panel overlays the per-layer distances with their across-layer median. Vertical lines indicate the displayed grasp, transport, and placement events. Large $d_t$ peaks are sparse and tend to occur near these transitions.}
    \label{fig:update_direction}
\end{figure*}

For an adapted layer, let $\Delta W_t = B_t A$ denote the effective low-rank update at control step $t$. We measure the change in update direction between adjacent steps by the Frobenius cosine distance
\begin{equation}
d_t =
1 -
\frac{
\langle \Delta W_t, \Delta W_{t-1} \rangle_F
}{
\|\Delta W_t\|_F \|\Delta W_{t-1}\|_F
}.
\label{eq:update_direction_distance}
\end{equation}
Figure~\ref{fig:update_direction} visualizes $d_t$ across adapted layers and timesteps for a representative rollout from LIBERO-Spatial. Direction changes are sparse over time and occur coherently across multiple adapted layers. We observe the same qualitative pattern in the other LIBERO suites; their corresponding visualizations are provided in Appendix~\ref{app:update_direction_all}. Across suites, peaks in $d_t$ tend to appear near salient manipulation events, such as grasp, transport transition, and placement, consistent with the role of descriptor-conditioned adaptation.

\paragraph{Quantitative alignment.}
Across evaluation episodes from the four LIBERO suites, we aggregate per-layer direction distances into a timestep-level update-variation score and correlate it with descriptor changes, $|\Delta \hat P_t|$ and $|\Delta \hat E_t|$. Table~\ref{tab:update_quant} first verifies that the learned descriptor-dependent components are
actively used: update-direction changes correlate with changes in the
predicted descriptors, especially the event/boundary descriptor. More importantly, high-\(d_t\) timesteps are enriched near gripper-transition proxies for grasping and placement, suggesting that the induced direction changes are aligned with salient control events rather than being uniform temporal noise.

\begin{table*}[t]
\centering
\caption{
Quantitative update-direction analysis across the four LIBERO suites.
For each suite, statistics are computed over 100 evaluation episodes. We report Spearman correlations between update-direction changes and descriptor changes over all valid timestep samples. High-\(d_t\) timesteps are defined as the top 10\% of timesteps within each suite, and transition neighborhoods are defined as \(\pm 5\) timesteps around automatically detected gripper-command transitions.
}
\label{tab:update_quant}
\small
\setlength{\tabcolsep}{5pt}
\begin{tabular}{llcccc}
\toprule
Analysis & Statistic
& Spatial
& Object
& Goal
& 10 \\
\midrule
\multirow{2}{*}{Descriptor alignment}
& Spearman $\rho(d_t, |\Delta \hat P_t|)$
& 0.502 & 0.439 & 0.497 & 0.440 \\
& Spearman $\rho(d_t, |\Delta \hat E_t|)$
& 0.712 & 0.743 & 0.709 & 0.737 \\
\midrule
\multirow{3}{*}{Transition enrichment}
& High-$d_t$ near transitions
& 21.0\% & 24.3\% & 21.8\% & 24.3\% \\
& High-$d_t$ outside transitions
& 8.1\% & 7.9\% & 8.7\% & 9.0\% \\
& Enrichment
& 2.60$\times$ & 3.10$\times$ & 2.52$\times$ & 2.71$\times$ \\
\bottomrule
\end{tabular}
\end{table*}

\section{Conclusion}
\label{sec:conclusion}

We introduced PhaseLoRA, a parameter-efficient fine-tuning method for continuous-action VLA policies that conditions LoRA updates on weakly supervised descriptors of fine-control tendency and event/boundary intensity. By modulating adaptation with trajectory-level control heterogeneity, PhaseLoRA improves over standard LoRA, matched-parameter High-rank LoRA, and structured LoRA variants on LIBERO. Ablations and update-direction analyses support the role of descriptor-conditioned update directions, and real-world tabletop experiments suggest that the trend can transfer to physical execution.

\paragraph{Limitations and scope.}
PhaseLoRA uses action-derived weak proxy descriptors rather than manual phase labels or
additional force/tactile sensing, which keeps the method lightweight but may require alternative
descriptors for task families where action statistics do not reflect fine-control or boundary events.
Our main comparisons cover the four LIBERO suites, while detailed ablations are mainly on
LIBERO-Spatial and real-world results are limited to four tabletop tasks on one robot platform.
\bibliographystyle{plainnat}
\bibliography{references}

@InProceedings{pmlr-v164-jang22a,
  title = 	 {BC-Z: Zero-Shot Task Generalization with Robotic Imitation Learning},
  author =       {Jang, Eric and Irpan, Alex and Khansari, Mohi and Kappler, Daniel and Ebert, Frederik and Lynch, Corey and Levine, Sergey and Finn, Chelsea},
  booktitle = 	 {Proceedings of the 5th Conference on Robot Learning},
  pages = 	 {991--1002},
  year = 	 {2022},
  editor = 	 {Faust, Aleksandra and Hsu, David and Neumann, Gerhard},
  volume = 	 {164},
  series = 	 {Proceedings of Machine Learning Research},
  month = 	 {08--11 Nov},
  publisher =    {PMLR},
}

@InProceedings{pmlr-v205-shridhar23a,
  title = 	 {Perceiver-Actor: A Multi-Task Transformer for Robotic Manipulation},
  author =       {Shridhar, Mohit and Manuelli, Lucas and Fox, Dieter},
  booktitle = 	 {Proceedings of The 6th Conference on Robot Learning},
  pages = 	 {785--799},
  year = 	 {2023},
  editor = 	 {Liu, Karen and Kulic, Dana and Ichnowski, Jeff},
  volume = 	 {205},
  series = 	 {Proceedings of Machine Learning Research},
  month = 	 {14--18 Dec},
  publisher =    {PMLR},
}

@inproceedings{BrohanBCCDFGHHH23,
  author = {Brohan, Anthony and Brown, Noah and Carbajal, Justice and Chebotar, Yevgen and Dabis, Joseph and Finn, Chelsea and Gopalakrishnan, Keerthana and Hausman, Karol and Herzog, Alexander and Hsu, Jasmine and Ibarz, Julian and Ichter, Brian and Irpan, Alex and Jackson, Tomas and Jesmonth, Sally and Joshi, Nikhil J. and Julian, Ryan and Kalashnikov, Dmitry and Kuang, Yuheng and Leal, Isabel and Lee, Kuang-Huei and Levine, Sergey and Lu, Yao and Malla, Utsav and Manjunath, Deeksha and Mordatch, Igor and Nachum, Ofir and Parada, Carolina and Peralta, Jodilyn and Perez, Emily and Pertsch, Karl and Quiambao, Jornell and Rao, Kanishka and Ryoo, Michael S. and Salazar, Grecia and Sanketi, Pannag R. and Sayed, Kevin and Singh, Jaspiar and Sontakke, Sumedh and Stone, Austin and Tan, Clayton and Tran, Huong T. and Vanhoucke, Vincent and Vega, Steve and Vuong, Quan and Xia, Fei and Xiao, Ted and Xu, Peng and Xu, Sichun and Yu, Tianhe and Zitkovich, Brianna},
  booktitle = {Robotics: Science and Systems},
  editor = {Bekris, Kostas E. and Hauser, Kris and Herbert, Sylvia L. and Yu, Jingjin},
  ee = {https://doi.org/10.15607/RSS.2023.XIX.025},
  isbn = {978-0-9923747-9-2},
  title = {{RT-1}: Robotics Transformer for Real-World Control at Scale.},
  year = 2023
}

@InProceedings{pmlr-v202-driess23a,
  title = 	 {{P}a{LM}-E: An Embodied Multimodal Language Model},
  author =       {Driess, Danny and Xia, Fei and Sajjadi, Mehdi S. M. and Lynch, Corey and Chowdhery, Aakanksha and Ichter, Brian and Wahid, Ayzaan and Tompson, Jonathan and Vuong, Quan and Yu, Tianhe and Huang, Wenlong and Chebotar, Yevgen and Sermanet, Pierre and Duckworth, Daniel and Levine, Sergey and Vanhoucke, Vincent and Hausman, Karol and Toussaint, Marc and Greff, Klaus and Zeng, Andy and Mordatch, Igor and Florence, Pete},
  booktitle = 	 {Proceedings of the 40th International Conference on Machine Learning},
  pages = 	 {8469--8488},
  year = 	 {2023},
  editor = 	 {Krause, Andreas and Brunskill, Emma and Cho, Kyunghyun and Engelhardt, Barbara and Sabato, Sivan and Scarlett, Jonathan},
  volume = 	 {202},
  series = 	 {Proceedings of Machine Learning Research},
  month = 	 {23--29 Jul},
  publisher =    {PMLR},
}

@InProceedings{pmlr-v229-zitkovich23a,
  title = 	 {{RT-2}: Vision-Language-Action Models Transfer Web Knowledge to Robotic Control},
  author =       {Zitkovich, Brianna and Yu, Tianhe and Xu, Sichun and Xu, Peng and Xiao, Ted and Xia, Fei and Wu, Jialin and Wohlhart, Paul and Welker, Stefan and Wahid, Ayzaan and Vuong, Quan and Vanhoucke, Vincent and Tran, Huong and Soricut, Radu and Singh, Anikait and Singh, Jaspiar and Sermanet, Pierre and Sanketi, Pannag R. and Salazar, Grecia and Ryoo, Michael S. and Reymann, Krista and Rao, Kanishka and Pertsch, Karl and Mordatch, Igor and Michalewski, Henryk and Lu, Yao and Levine, Sergey and Lee, Lisa and Lee, Tsang-Wei Edward and Leal, Isabel and Kuang, Yuheng and Kalashnikov, Dmitry and Julian, Ryan and Joshi, Nikhil J. and Irpan, Alex and Ichter, Brian and Hsu, Jasmine and Herzog, Alexander and Hausman, Karol and Gopalakrishnan, Keerthana and Fu, Chuyuan and Florence, Pete and Finn, Chelsea and Dubey, Kumar Avinava and Driess, Danny and Ding, Tianli and Choromanski, Krzysztof Marcin and Chen, Xi and Chebotar, Yevgen and Carbajal, Justice and Brown, Noah and Brohan, Anthony and Arenas, Montserrat Gonzalez and Han, Kehang},
  booktitle = 	 {Proceedings of The 7th Conference on Robot Learning},
  pages = 	 {2165--2183},
  year = 	 {2023},
  editor = 	 {Tan, Jie and Toussaint, Marc and Darvish, Kourosh},
  volume = 	 {229},
  series = 	 {Proceedings of Machine Learning Research},
  month = 	 {06--09 Nov},
  publisher =    {PMLR},
}

@inproceedings{openxembodiment2024,
  title={Open {X-E}mbodiment: Robotic Learning Datasets and {RT-X} Models},
  author={{Open X-Embodiment Collaboration}},
  booktitle={IEEE International Conference on Robotics and Automation (ICRA)},
  year={2024},
  pages={6892-6903},
}

@InProceedings{pmlr-v270-kim25c,
  title = 	 {{OpenVLA}: An Open-Source Vision-Language-Action Model},
  author =       {Kim, Moo Jin and Pertsch, Karl and Karamcheti, Siddharth and Xiao, Ted and Balakrishna, Ashwin and Nair, Suraj and Rafailov, Rafael and Foster, Ethan P and Sanketi, Pannag R and Vuong, Quan and Kollar, Thomas and Burchfiel, Benjamin and Tedrake, Russ and Sadigh, Dorsa and Levine, Sergey and Liang, Percy and Finn, Chelsea},
  booktitle = 	 {Proceedings of The 8th Conference on Robot Learning},
  pages = 	 {2679--2713},
  year = 	 {2025},
  editor = 	 {Agrawal, Pulkit and Kroemer, Oliver and Burgard, Wolfram},
  volume = 	 {270},
  series = 	 {Proceedings of Machine Learning Research},
  month = 	 {06--09 Nov},
  publisher =    {PMLR},
}

@InProceedings{pmlr-v305-black25a,
  title = 	 {$\pi_{0.5}$: a Vision-Language-Action Model with Open-World Generalization},
  author =       {Black, Kevin and Brown, Noah and Darpinian, James and Dhabalia, Karan and Driess, Danny and Esmail, Adnan and Equi, Michael Robert and Finn, Chelsea and Fusai, Niccolo and Galliker, Manuel Y. and Ghosh, Dibya and Groom, Lachy and Hausman, Karol and ichter, brian and Jakubczak, Szymon and Jones, Tim and Ke, Liyiming and LeBlanc, Devin and Levine, Sergey and Li-Bell, Adrian and Mothukuri, Mohith and Nair, Suraj and Pertsch, Karl and Ren, Allen Z. and Shi, Lucy Xiaoyang and Smith, Laura and Springenberg, Jost Tobias and Stachowicz, Kyle and Tanner, James and Vuong, Quan and Walke, Homer and Walling, Anna and Wang, Haohuan and Yu, Lili and Zhilinsky, Ury},
  booktitle = 	 {Proceedings of The 9th Conference on Robot Learning},
  pages = 	 {17--40},
  year = 	 {2025},
  editor = 	 {Lim, Joseph and Song, Shuran and Park, Hae-Won},
  volume = 	 {305},
  series = 	 {Proceedings of Machine Learning Research},
  month = 	 {27--30 Sep},
  publisher =    {PMLR},
}

@article{Liu2024RDT1BAD,
  title={{RDT-1B}: a Diffusion Foundation Model for Bimanual Manipulation},
  author={Songming Liu and Lingxuan Wu and Bangguo Li and Hengkai Tan and Huayu Chen and Zhengyi Wang and Ke Xu and Hang Su and Jun Zhu},
  journal={ArXiv},
  year={2024},
  volume={abs/2410.07864},
}

@article{Kim2025FineTuningVM,
  title={Fine-Tuning Vision-Language-Action Models: Optimizing Speed and Success},
  author={Kim, Moo Jin and Finn, Chelsea and Liang, Percy},
  journal={arXiv preprint arXiv:2502.19645},
  year={2025}
}

@inproceedings{kim2026adaptive,
  title     = {Adaptive Capacity Allocation for Vision Language Action Fine-tuning},
  author    = {Kim, Donghoon and Bae, Minji and Nam, Unghui and Kim, Gyeonghun and Lee, Suyun and Shim, Kyuhong and Shim, Byonghyo},
  booktitle = {Proceedings of the IEEE International Conference on Robotics and Automation (ICRA)},
  year      = {2026},
}

@article{hu2026simplerecipeworksvisionlanguageaction,
  title   = {Simple Recipe Works: Vision-Language-Action Models are Natural Continual Learners with Reinforcement Learning},
  author  = {Hu, Jiaheng and Shim, Jay and Tang, Chen and Sung, Yoonchang and Liu, Bo and Stone, Peter and Martin-Martin, Roberto},
  journal = {arXiv preprint arXiv:2603.11653},
  year    = {2026},
}

@article{huang2025corft,
  title={CO-RFT: Efficient Fine-Tuning of Vision-Language-Action Models through Chunked Offline Reinforcement Learning},
  author={Huang, Dongchi and Fang, Zhirui and Zhang, Tianle and Li, Yihang and Zhao, Lin and Xia, Chunhe},
  journal={arXiv preprint arXiv:2508.02219},
  year={2025},
}

@misc{li2025vlarftvisionlanguageactionreinforcementfinetuning,
  title        = {{VLA-RFT}: Vision-Language-Action Reinforcement Fine-tuning with Verified Rewards in World Simulators},
  author       = {Li, Hengtao and Ding, Pengxiang and Suo, Runze and Wang, Yihao and Ge, Zirui and Zang, Dongyuan and Yu, Kexian and Sun, Mingyang and Zhang, Hongyin and Wang, Donglin and Su, Weihua},
  year         = {2025},
  eprint       = {2510.00406},
  archivePrefix = {arXiv},
  primaryClass = {cs.RO},
  doi          = {10.48550/arXiv.2510.00406}
}

@inproceedings{valipour-etal-2023-dylora,
    title = "{D}y{L}o{RA}: Parameter-Efficient Tuning of Pre-trained Models using Dynamic Search-Free Low-Rank Adaptation",
    author = "Valipour, Mojtaba  and
      Rezagholizadeh, Mehdi  and
      Kobyzev, Ivan  and
      Ghodsi, Ali",
    editor = "Vlachos, Andreas  and
      Augenstein, Isabelle",
    booktitle = "Proceedings of the 17th Conference of the European Chapter of the Association for Computational Linguistics",
    month = may,
    year = "2023",
    address = "Dubrovnik, Croatia",
    publisher = "Association for Computational Linguistics",
    doi = "10.18653/v1/2023.eacl-main.239",
    pages = "3274--3287",
}

@inproceedings{zhang2023adalora,
  title={AdaLoRA: Adaptive Budget Allocation for Parameter-Efficient Fine-Tuning},
  author={Zhang, Qingru and Chen, Minshuo and Bukharin, Alexander and He, Pengcheng and Cheng, Yu and Chen, Weizhu and Zhao, Tuo},
  booktitle={The Eleventh International Conference on Learning Representations},
  year={2023},
}

@InProceedings{pmlr-v235-liu24bn,
  title = 	 {{D}o{RA}: Weight-Decomposed Low-Rank Adaptation},
  author =       {Liu, Shih-Yang and Wang, Chien-Yi and Yin, Hongxu and Molchanov, Pavlo and Wang, Yu-Chiang Frank and Cheng, Kwang-Ting and Chen, Min-Hung},
  booktitle = 	 {Proceedings of the 41st International Conference on Machine Learning},
  pages = 	 {32100--32121},
  year = 	 {2024},
  editor = 	 {Salakhutdinov, Ruslan and Kolter, Zico and Heller, Katherine and Weller, Adrian and Oliver, Nuria and Scarlett, Jonathan and Berkenkamp, Felix},
  volume = 	 {235},
  series = 	 {Proceedings of Machine Learning Research},
  month = 	 {21--27 Jul},
  publisher =    {PMLR},
}

@article{buehler2024xlora_journal,
    author = {Buehler, Eric L. and Buehler, Markus J.},
    title = "{X-LoRA: Mixture of low-rank adapter experts, a flexible framework for large language models with applications in protein mechanics and molecular design}",
    journal = {APL Machine Learning},
    volume = {2},
    number = {2},
    pages = {026119},
    year = {2024},
    month = {05},
    issn = {2770-9019},
    doi = {10.1063/5.0203126},
    eprint = {https://pubs.aip.org/aip/aml/article-pdf/doi/10.1063/5.0203126/19964043/026119\_1\_5.0203126.pdf},
}

@inproceedings{pmlr-v267-ge25d,
  title     = {Mixture of {LoRA} Experts},
  author    = {Wu, Xun and Huang, Shaohan and Wei, Furu},
  booktitle = {The Twelfth International Conference on Learning Representations},
  year      = {2024},
}

@inproceedings{ivison-etal-2023-hint,
    title = "{HINT}: Hypernetwork Instruction Tuning for Efficient Zero- and Few-Shot Generalisation",
    author = "Ivison, Hamish  and
      Bhagia, Akshita  and
      Wang, Yizhong  and
      Hajishirzi, Hannaneh  and
      Peters, Matthew",
    editor = "Rogers, Anna  and
      Boyd-Graber, Jordan  and
      Okazaki, Naoaki",
    booktitle = "Proceedings of the 61st Annual Meeting of the Association for Computational Linguistics (Volume 1: Long Papers)",
    month = jul,
    year = "2023",
    address = "Toronto, Canada",
    publisher = "Association for Computational Linguistics",
    doi = "10.18653/v1/2023.acl-long.631",
    pages = "11272--11288",
}

@inproceedings{lv-etal-2024-hyperlora,
    title = "{H}yper{L}o{RA}: Efficient Cross-task Generalization via Constrained Low-Rank Adapters Generation",
    author = "Lv, Chuancheng  and
      Li, Lei  and
      Zhang, Shitou  and
      Chen, Gang  and
      Qi, Fanchao  and
      Zhang, Ningyu  and
      Zheng, Hai-Tao",
    editor = "Al-Onaizan, Yaser  and
      Bansal, Mohit  and
      Chen, Yun-Nung",
    booktitle = "Findings of the Association for Computational Linguistics: EMNLP 2024",
    month = nov,
    year = "2024",
    address = "Miami, Florida, USA",
    publisher = "Association for Computational Linguistics",
    doi = "10.18653/v1/2024.findings-emnlp.956",
    pages = "16376--16393",
}

@article{jin2024conditionalloraparametergeneration,
  title   = {Conditional LoRA Parameter Generation},
  author  = {Jin, Xiaolong and Wang, Kai and Tang, Dongwen and Zhao, Wangbo and Zhou, Yukun and Tang, Junshu and You, Yang},
  journal = {arXiv preprint arXiv:2408.01415},
  year    = {2024},
}

@ARTICLE{vonhartz2024artimitationlearninglonghorizon,
  author={von Hartz, Jan Ole and Welschehold, Tim and Valada, Abhinav and Boedecker, Joschka},
  journal={IEEE Robotics and Automation Letters}, 
  title={The Art of Imitation: Learning Long-Horizon Manipulation Tasks From Few Demonstrations}, 
  year={2024},
  volume={9},
  number={12},
  pages={11369-11376},
  doi={10.1109/LRA.2024.3487506}}

@misc{cheng2026tacumimultimodaluniversalmanipulation,
      title={{TacUMI}: A Multi-Modal Universal Manipulation Interface for Contact-Rich Tasks}, 
      author={Tailai Cheng and Kejia Chen and Lingyun Chen and Liding Zhang and Yue Zhang and Yao Ling and Mahdi Hamad and Zhenshan Bing and Fan Wu and Karan Sharma and Alois Knoll},
      year={2026},
      eprint={2601.14550},
      archivePrefix={arXiv},
      primaryClass={cs.RO},
}

@inproceedings{li2026robointerholisticintermediaterepresentation,
  title={RoboInter: A Holistic Intermediate Representation Suite Towards Robotic Manipulation},
  author={Li, Hao and Wang, Ziqin and Ding, Zi-han and Yang, Shuai and Chen, Yilun and Tian, Yang and Hu, Xiaolin and Wang, Tai and Lin, Dahua and Zhao, Feng and Liu, Si and Pang, Jiangmiao},
  booktitle={International Conference on Learning Representations},
  year={2026},
}

@article{zhao2025touch,
  title={Touch begins where vision ends: Generalizable policies for contact-rich manipulation},
  author={Zhao, Zifan and Haldar, Siddhant and Cui, Jinda and Pinto, Lerrel and Bhirangi, Raunaq M.},
  journal={arXiv preprint arXiv:2506.13762},
  year={2025}
}

@article{doi:10.55092/rl20260001,
author = {Hao, Peng and Zhang, Chaofan and Li, Dingzhe and Cao, Xiaoge and Hao, Xiaoshuai and Cui, Shaowei and Wang, Shuo},
title = {{TLA}: tactile-language-action model for contact-rich manipulation},
journal = {Robot Learning},
volume = {3},
number = {1},
year = {2026},
doi = {10.55092/rl20260001},
}

@inproceedings{yu2025forcevlaenhancingvlamodels,
  title     = {{ForceVLA}: Enhancing {VLA} Models with a Force-aware {MoE} for Contact-rich Manipulation},
  author    = {Yu, Jiawen and Liu, Hairuo and Yu, Qiaojun and Ren, Jieji and Hao, Ce and Ding, Haitong and Huang, Guangyu and Huang, Guofan and Song, Yan and Cai, Panpan and Zhang, Wenqiang and Lu, Cewu},
  booktitle = {The Thirty-ninth Annual Conference on Neural Information Processing Systems},
  year      = {2025},
}

@article{huang2025tactilevlaunlockingvisionlanguageactionmodels,
  title   = {{Tactile-VLA}: Unlocking Vision-Language-Action Model's Physical Knowledge for Tactile Generalization},
  author  = {Huang, Jialei and Wang, Shuo and Lin, Fanqi and Hu, Yihang and Wen, Chuan and Gao, Yang},
  journal = {arXiv preprint arXiv:2507.09160},
  year    = {2025},
  doi     = {10.48550/arXiv.2507.09160}
}

@misc{gubernatorov2026hapticvlacontactrichmanipulationvisionlanguageaction,
      title={{HapticVLA}: Contact-Rich Manipulation via Vision-Language-Action Model without Inference-Time Tactile Sensing}, 
      author={Konstantin Gubernatorov and Mikhail Sannikov and Ilya Mikhalchuk and Egor Kuznetsov and Makar Artemov and Ogunwoye Faith Ouwatobi and Marcelino Fernando and Artem Asanov and Ziang Guo and Dzmitry Tsetserukou},
      year={2026},
      eprint={2603.15257},
      archivePrefix={arXiv},
      primaryClass={cs.RO},
}

@inproceedings{wei2025moka,
  title={MokA: Multimodal Low-Rank Adaptation for MLLMs},
  author={Wei, Yake and Miao, Yu and Zhou, Dongzhan and Hu, Di},
  booktitle={Advances in Neural Information Processing Systems},
  year={2025}
}

@inproceedings{ghosh2024octo,
    title={Octo: An Open-Source Generalist Robot Policy},
    author = {{Octo Model Team} and Dibya Ghosh and Homer Walke and Karl Pertsch and Kevin Black and Oier Mees and Sudeep Dasari and Joey Hejna and Charles Xu and Jianlan Luo and Tobias Kreiman and {You Liang} Tan and Lawrence Yunliang Chen and Pannag Sanketi and Quan Vuong and Ted Xiao and Dorsa Sadigh and Chelsea Finn and Sergey Levine},
    booktitle = {Proceedings of Robotics: Science and Systems},
    address  = {Delft, Netherlands},
    year = {2024},
}

@inproceedings{hu2022lowrank,
  title     = {{LoRA}: Low-Rank Adaptation of Large Language Models},
  author    = {Hu, Edward J. and Shen, Yelong and Wallis, Phillip and Allen-Zhu, Zeyuan and Li, Yuanzhi and Wang, Shean and Wang, Lu and Chen, Weizhu},
  booktitle = {International Conference on Learning Representations},
  year      = {2022},
  publisher = {OpenReview.net},
}

@inproceedings{liu2024libero,
author = {Liu, Bo and Zhu, Yifeng and Gao, Chongkai and Feng, Yihao and Liu, Qiang and Zhu, Yuke and Stone, Peter},
title = {{LIBERO}: benchmarking knowledge transfer for lifelong robot learning},
year = {2023},
publisher = {Curran Associates Inc.},
address = {Red Hook, NY, USA},
booktitle = {Proceedings of the 37th International Conference on Neural Information Processing Systems},
articleno = {1939},
numpages = {16},
location = {New Orleans, LA, USA},
series = {NIPS '23}
}

@article{cadene2026lerobot,
  title={{LeRobot}: An Open-Source Library for End-to-End Robot Learning},
  author={Cadene, Remi and Aliberts, Simon and Capuano, Francesco and others},
  journal={arXiv preprint arXiv:2602.22818},
  year={2026}
}

@misc{hu2026behaviorvla,
  title         = {From Abstraction to Instantiation: Learning Behavioral
                   Representation for {Vision-Language-Action} Model},
  author        = {Hu, Bing and Li, Zaijing and Shao, Rui and Chen, Junda
                   and Liu, April Hua and Zheng, Wei-Shi and Nie, Liqiang},
  year          = {2026},
  eprint        = {2605.22671},
  archivePrefix = {arXiv},
  primaryClass  = {cs.CV},
  url           = {https://arxiv.org/abs/2605.22671}
}

@misc{wang2026magvla,
  title         = {{Mag-VLA}: {Vision-Language-Action} Model for Bimanual
                   Magnetically Actuated Microrobot Manipulation},
  author        = {Wang, Yongchen and Lu, Kangyi and Wei, Lan and Zhang, Dandan},
  year          = {2026},
  eprint        = {2605.28486},
  archivePrefix = {arXiv},
  primaryClass  = {cs.RO},
  url           = {https://arxiv.org/abs/2605.28486}
}

%%%%%%%%%%%%%%%%%%%%%%%%%%%%%%%%%%%%%%%%%%%%%%%%%%%%%%%%%%%%

\appendix

\section{Implementation and Reproducibility Details}
\label{app:implementation}

This section provides additional implementation details for reproducing the LIBERO experiments.

\subsection{Backbone and action representation}
\label{app:backbone}

All experiments are built on the $\pi_{0.5}$ VLA backbone. We initialize the model from the public base checkpoint at
\texttt{gs://openpi-assets/checkpoints/pi05\_base/params}. The VLM component uses the
\texttt{gemma\_2b\_lora} variant, and the action expert uses the
\texttt{gemma\_300m\_lora} variant.

Images are resized to $224 \times 224$ during both training and evaluation. For LIBERO evaluation,
the policy receives two real RGB camera views, \texttt{base\_0\_rgb} and
\texttt{left\_wrist\_0\_rgb}. The third image slot, \texttt{right\_wrist\_0\_rgb}, is filled with zeros
as a dummy view for compatibility with the model input format. Language instructions are taken
directly from the LIBERO task language annotations.

The model predicts action chunks with action dimension 32 and horizon $H=10$. For LIBERO control,
we use the first 7 action dimensions, corresponding to 3 translational dimensions, 3 rotational
dimensions, and 1 gripper command. PhaseLoRA itself does not require a particular rotation parameterization.
In our LIBERO implementation, however, the first seven action dimensions are
\[
    \mathbf{a}_t =
    \left[
        \Delta x_t,\,
        \Delta y_t,\,
        \Delta z_t,\,
        \Delta r_t^x,\,
        \Delta r_t^y,\,
        \Delta r_t^z,\,
        g_t
    \right] \in [-1,1]^7.
\]
The first six dimensions follow the normalized delta-command convention of
the robosuite v1.4.1 \texttt{OSC\_POSE} controller. The first three components
specify relative Cartesian-position commands and are scaled by the controller
to a maximum absolute value of \(0.05\,\mathrm{m}\) per dimension. The next
three components specify a relative axis--angle rotation vector and are scaled
to a maximum absolute value of \(0.5\,\mathrm{rad}\) per dimension. The gripper
command \(g_t\) lies in \([-1,1]\), where \(-1\) denotes opening and \(+1\)
denotes closing. LIBERO actions are already represented as delta commands, so
we do not apply an additional delta-action transformation. During training,
actions are normalized per dimension using training-set quantile statistics.
At inference time, the predicted actions are inverse-normalized before their
first seven dimensions are passed to the environment. During closed-loop evaluation, the policy predicts a 10-step action chunk but executes only the first 5 low-level actions before replanning. Thus, the default replanning interval is 5 control steps.

\subsection{LIBERO benchmark and evaluation protocol}

We evaluate on the four standard LIBERO suites used in the main paper:
LIBERO-Spatial, LIBERO-Object, LIBERO-Goal, and LIBERO-10. We use the official LIBERO datasets
and task language annotations from the LIBERO benchmark. Each suite-specific policy is fine-tuned on
the corresponding LIBERO training split and evaluated on the corresponding suite.

Unless otherwise stated, evaluation uses 10 trials per task and 3 seeds. The success
rate for a suite is computed as the episode-level success rate,
\[
    \mathrm{SuccessRate}
    =
    \frac{\text{\# successful evaluation episodes}}
         {\text{\# total evaluation episodes}}.
\]
The average reported in the main table is the average across the four LIBERO suite scores.

\subsection{LoRA and PhaseLoRA configuration}

Our implementation follows the LoRA implementation in the Gemma modules of the $\pi_{0.5}$
codebase. Instead of specifying a separate PEFT-style list of target modules, the attention and
feed-forward linear projections in the Gemma implementation are wrapped with LoRA modules.
Specifically, LoRA is applied to the attention projections, including the query/key/value and output
projections, as well as to the feed-forward gate and linear projections. We do not attach a separate
LoRA module to an additional action head.

For standard LoRA and PhaseLoRA, the VLM LoRA rank is 48 with scaling factor
$\alpha=48$, and the action-expert LoRA rank is 96 with scaling factor $\alpha=96$. For the
matched-parameter High-rank LoRA baseline, the VLM LoRA rank is 80 with $\alpha=80$, and the
action-expert LoRA rank is 160 with $\alpha=160$. The reported trainable-parameter percentages
are measured relative to the full $\pi_{0.5}$ backbone parameter count. The LoRA implementation
does not use LoRA dropout.

For PhaseLoRA, each adapted layer uses four left factors and one
shared right factor. Given router-predicted descriptors $(\hat P_t,\hat E_t)$ at control step $t$, the
effective left factor is
\[
    B_t = B_0 + \hat P_t B_P + \hat E_t B_E + \hat P_t \hat E_t B_{PE},
\]
and the low-rank update is
\[
    \Delta W_t = B_t A.
\]
Thus, for a fixed pair $(\hat P_t,\hat E_t)$, the update remains low-rank, while its effective left
factor can vary across control steps. The shared LoRA component follows the standard LoRA
initialization: the right factor is initialized with small Gaussian noise and the corresponding left
factor is initialized to zero. The descriptor-dependent left factors $B_P$, $B_E$, and $B_{PE}$ are
initialized to zero, so the model starts from an ordinary LoRA update and learns descriptor-dependent
deviations only during fine-tuning.

\subsection{Router architecture}

The phase router predicts two scalar descriptors at each control step: a fine-control descriptor
$\hat P_t$ and an event/boundary descriptor $\hat E_t$. The router uses three input branches:
a visual-language prefix summary, recent action and gripper history.
The visual-language summary is computed by masked mean pooling over the multimodal prefix
embeddings. For the $\texttt{gemma\_2b\_lora}$ VLM used in our experiments, this prefix summary has
dimension 2048. The recent-action branch uses
the previous 6 action chunks, with the first 5 executed actions retained from each chunk, yielding
a history window of 30 low-level action steps.

For each low-level action step in the history window, the router forms a 23-dimensional feature:
7 action dimensions, 7 first-order differences, 7 second-order differences, 1 gripper state feature,
and 1 gripper-change feature. The prefix summary and action-history features are each projected to the router hidden dimension of 128. The temporal action-history branch is then
processed by a single-layer, unidirectional GRU with hidden dimension 128. The fused router hidden
state is passed to two linear heads, followed by sigmoid activations, to produce
$\hat P_t,\hat E_t \in [0,1]$.

At the beginning of an episode, when fewer than 30 previous low-level actions are available, the
history window is left-padded with zeros. If no history is available, the router receives an all-zero
history window. During training, the router uses offline action history to avoid leakage from the
current target action chunk. During evaluation, the action history is updated online from the actions
previously executed by the policy.

\subsection{Weak descriptor targets}

The weak descriptor targets are precomputed offline and injected during training. The fine-control
target is derived from a robustly normalized motion statistic over the trajectory. In the implementation,
the translational action magnitude is combined with the rotational action magnitude using rotation
weight 0.5 and with the gripper-command change using gripper weight 0.1. The event/boundary
target is derived from action jerk.

All trajectory statistics are robustly normalized with median and median absolute deviation, using
$\epsilon=10^{-6}$ for numerical stability. The sigmoid inputs are clipped to $[-60,60]$ before applying
the sigmoid, and the resulting targets are clamped to $[0,1]$.
The gripper command is treated as a continuous command rather than a binary open/close label.

\subsection{Training objective and optimization}

The main policy objective follows the $\pi_{0.5}$ training objective. The policy is trained with a
flow-matching mean-squared error objective, where the target velocity is defined as the difference
between sampled noise and the action target. The loss is applied to the predicted action tensor as a
whole; we do not use separate manually tuned weights for translation, rotation, and gripper dimensions.

The router is supervised with the weak descriptor targets using
\[
    \mathcal{L}_{\mathrm{router}}
    =
    0.5 \, \mathrm{Huber}(\hat P_t, P_t)
    +
    0.2 \, \mathrm{BCE}(\hat E_t, E_t),
\]
where the Huber delta for the fine-control target is 0.1.

All models are trained with AdamW. We use effectively no weight decay, setting it to $10^{-10}$, and apply gradient clipping with maximum norm 1.0. For the descriptor-dependent LoRA regularizer in Eq.~(12), we set
\[
\lambda^{\mathrm{reg}}_{P}
=
\lambda^{\mathrm{reg}}_{E}
=
\lambda^{\mathrm{reg}}_{PE}
=
10^{-5}
\]
for all LIBERO and real-world experiments. The learning rate is linearly warmed up for 10,000 steps to $5\times 10^{-5}$ and then kept constant for the remainder of training. We use batch size 32 and train each run for 30,000 optimization steps. Training uses bfloat16 precision. Exponential moving average is disabled. The training configuration resizes images to $224 \times 224$; in the PyTorch training path, non-wrist images additionally use a small random crop followed by resizing back to the target resolution. We do not use checkpoint averaging or early stopping.

\subsection{Baseline fairness}

For LoRA-based methods, we keep the backbone, data split, image preprocessing, action
representation, training objective, batch size, learning-rate schedule, number of training steps, and
adapted module family fixed. Standard LoRA and PhaseLoRA use the same base LoRA
ranks, while High-rank LoRA increases the ranks to approximately match the trainable-parameter
budget of PhaseLoRA. LoRA-SP is applied to the same family of Gemma attention and
feed-forward modules. Its key settings are energy threshold 0.9, router hidden dimension 256,
spectral loss weight $10^{-2}$, router loss weight $10^{-3}$, and inference-time pruning enabled.

\subsection{Compute resources}
\label{app:compute}
All LIBERO experiments were run on NVIDIA RTX 5090 GPUs. Each training run used a single GPU with bfloat16 precision and batch size 32. The peak GPU memory usage of PhaseLoRA was approximately \(22106\,\mathrm{MiB}\) per run. A single suite-specific LIBERO training run of 30,000 optimization steps took approximately 24 GPU-hours, depending on the method and suite.

For the PhaseLoRA main result, one seed across the four LIBERO suites required approximately \(4 \times 24 = 96\) GPU-hours, and the three-seed PhaseLoRA result required approximately \(288\) GPU-hours. The full main LIBERO comparison in Table~\ref{tab:libero_main} includes six methods, four suites, and three seeds, requiring approximately
\[
6 \times 4 \times 3 \times 24 = 1728
\]
GPU-hours. The additional LIBERO-Spatial ablations in Table~\ref{tab:ablations} required approximately
\[
6 \times 1 \times 3 \times 24 = 432
\]
GPU-hours. Thus, the reported LIBERO training runs required approximately \(2160\) GPU-hours in total.

The experiments used approximately 100GB of local SSD storage for the LIBERO datasets, the \(\pi0.5\) pretrained checkpoint, training checkpoints, logs, and cached artifacts. Real-world evaluation was run on the robot platform and did not require additional large-scale model training beyond the fine-tuned checkpoints. Preliminary debugging runs and failed experiments required additional compute, but they are not included in the totals above because they were not part of the final reported experimental protocol.

\section{Real-world Benchmark Details}
\label{app:real_world_details}

This section provides additional details for the real-world manipulation benchmark used in Sec.~\ref{subsec:real_main}.
The real-world experiments are intended as supporting evidence for the controlled LIBERO results,
rather than as a claim of broad real-world generalization across robot platforms or task families.

\paragraph{Robot platform.}
We use an AgileX Piper robotic arm equipped with a two-finger gripper. The robot is mounted on a
tabletop workspace and evaluated in closed-loop execution. All real-world policies are initialized from
the same pretrained VLA backbone as the LIBERO experiments and are fine-tuned using the same
method-specific adapter configuration as in the simulation benchmark, unless otherwise stated.

\begin{figure}[t]
    \centering
    \begin{subfigure}[t]{0.22\linewidth}
        \centering
        \includegraphics[width=\linewidth]{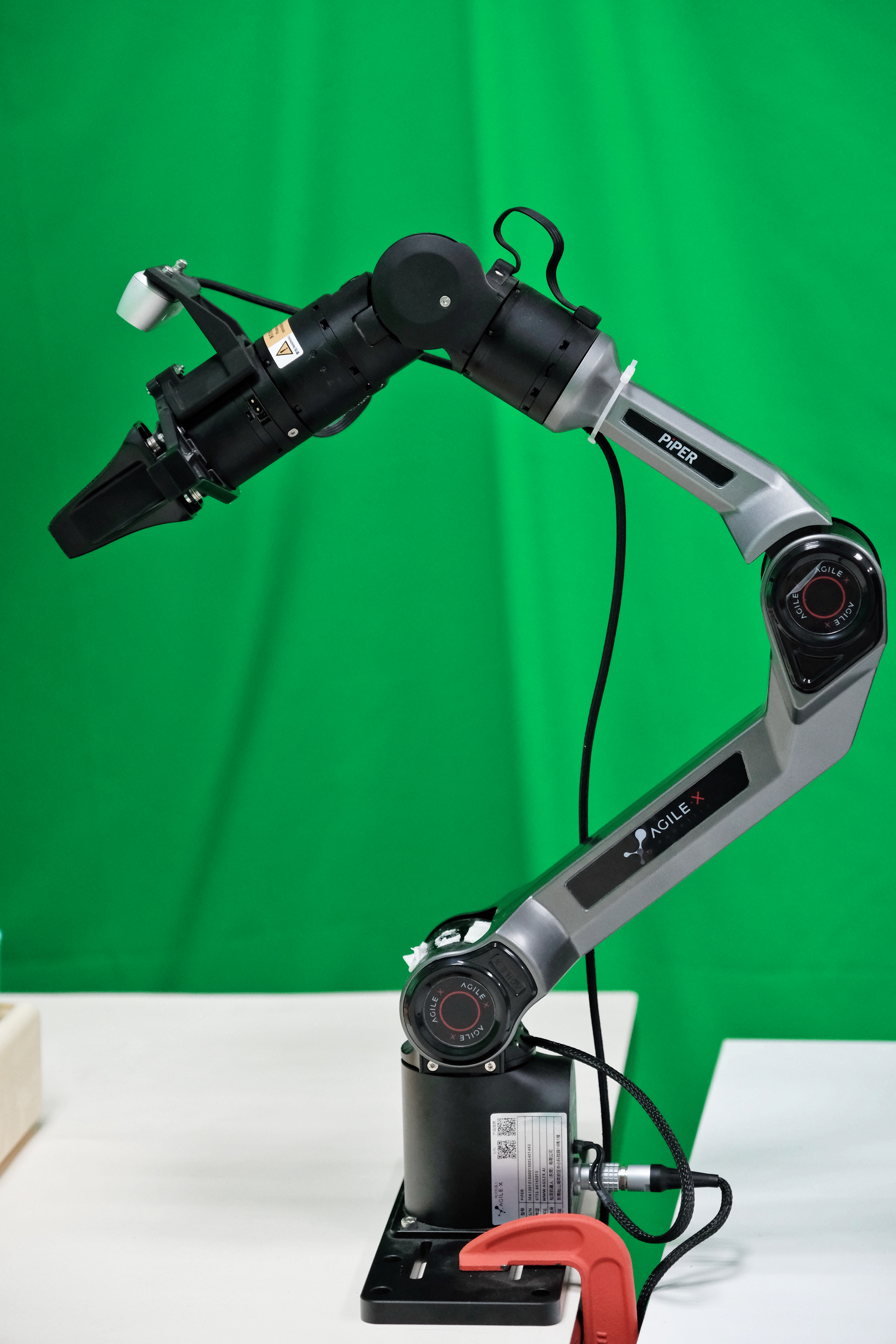}
        \caption{Piper arm with wrist camera.}
    \end{subfigure}
    \hfill
    \begin{subfigure}[t]{0.22\linewidth}
        \centering
        \includegraphics[width=\linewidth]{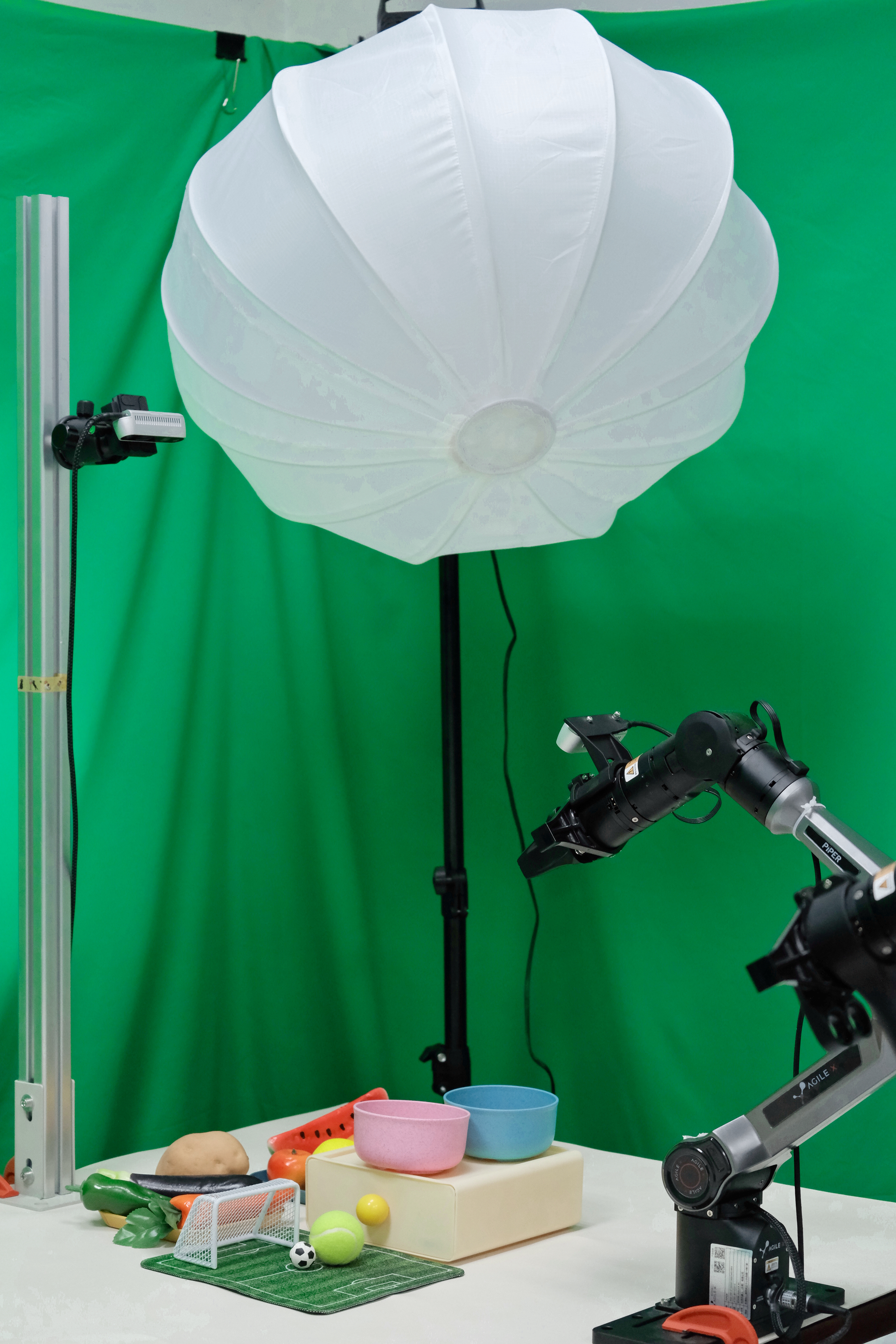}
        \caption{Full tabletop setup.}
    \end{subfigure}
    \hfill
    \begin{subfigure}[t]{0.50\linewidth}
        \centering
        \includegraphics[width=\linewidth]{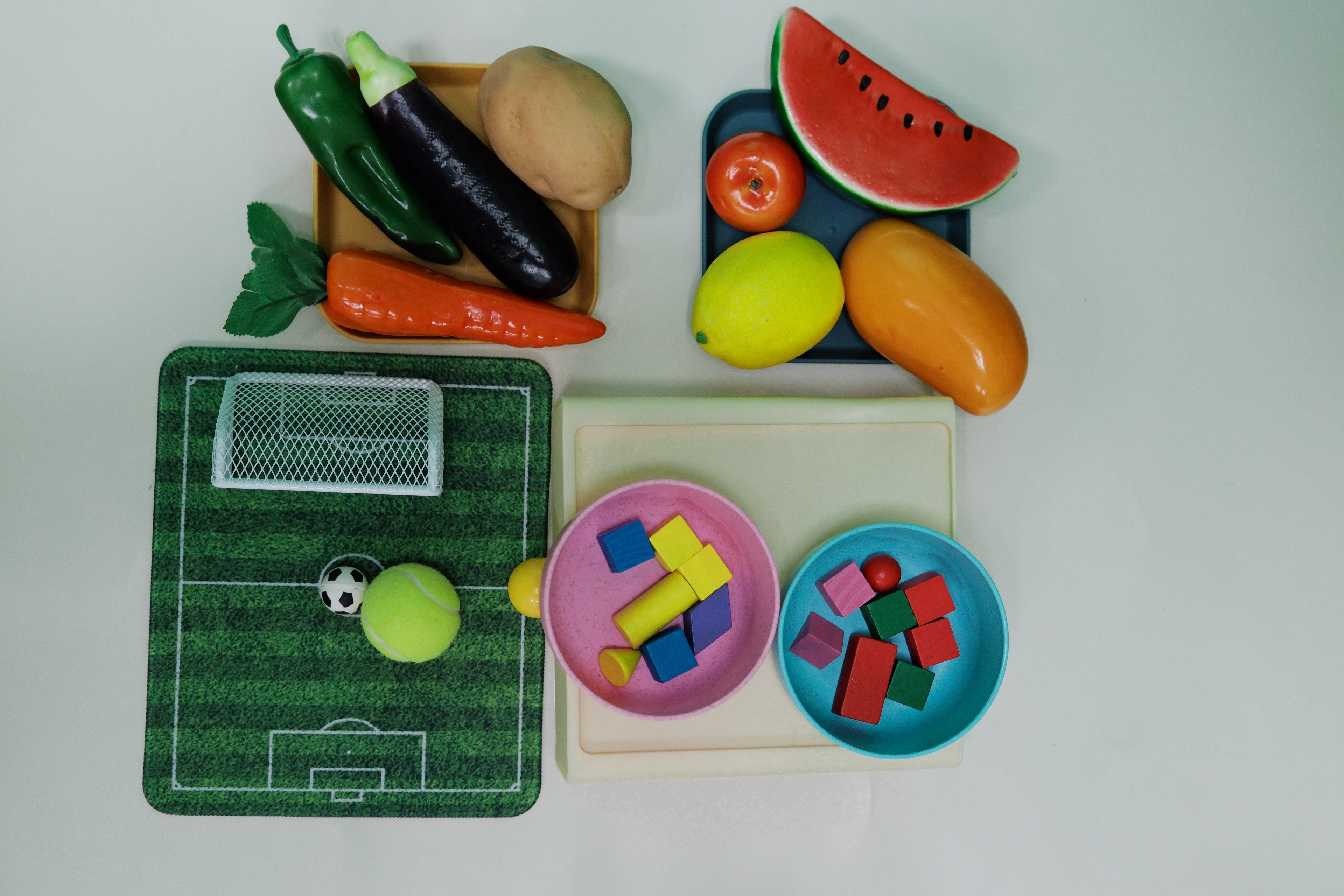}
        \caption{External-camera view.}
    \end{subfigure}
    \caption{
    Real-world benchmark setup.
    (a) Piper robotic arm with a two-finger gripper and a wrist-mounted RGB camera.
    (b) Full tabletop setup with a fixed external RGB camera, lighting, task objects, and the robot workspace.
    (c) Example external-camera observation showing the plates, objects, and distractors used in evaluation.
    The policy uses two RGB views and language instructions, without proprioceptive, force, or tactile inputs.
    }
    \label{fig:real_world_setup}
\end{figure}

\paragraph{Observation setup.}
The policy receives RGB observations from two cameras: a wrist-mounted RGB camera near the
gripper and a fixed external RGB camera observing the tabletop workspace. Example views of the
hardware and workspace are shown in Fig.~\ref{fig:real_world_setup}. All RGB images are resized to
$224 \times 224$ before being passed to the VLA policy. The policy uses only RGB observations and
the language instruction. We do not provide proprioceptive state, force sensing, or tactile sensing to
the policy during inference.

\paragraph{Tasks.}
The benchmark contains four tabletop manipulation tasks:
moving a tennis ball from a yellow plate to a blue plate, removing a cuboid from a blue plate,
picking a red cube into a yellow plate, and putting an apple on a yellow plate. These tasks require
grasping, transport, object release, and placement under real robot execution. Each task contains
100 demonstrations for fine-tuning. Demonstrations are collected using the same camera setup and
robot control interface used during policy evaluation.

\paragraph{Action chunking and closed-loop control.}
For real-world evaluation, the policy predicts action chunks with horizon $H=50$. During closed-loop
execution, we execute the first 10 low-level actions from each predicted chunk and then replan.
Thus, each policy query produces a 50-step action chunk, while the robot is controlled in a receding
horizon manner with a replanning interval of 10 executed low-level control steps. This chunking
setup is kept fixed across all compared methods in the real-world benchmark.

\paragraph{Training and baseline fairness.}
For each method, the policy is fine-tuned on the same set of demonstrations for the corresponding
task. The compared methods use the same pretrained backbone, observation inputs, language
instructions, training data, action chunking setting, optimizer, batch size, learning-rate schedule, and
number of training steps. The High-rank LoRA baseline is chosen to approximately match the
trainable-parameter budget of PhaseLoRA, so the real-world comparison tests whether the observed
trend is explained solely by increased adaptation capacity.

\paragraph{Evaluation protocol.}
For each method and task, we evaluate 10 physical trials for each of 3 random seeds, resulting in
30 trials per task per method. Success is judged manually according to pre-defined task-specific
criteria. An episode is counted as successful if the target object is moved to or removed from the
specified target region by the end of the episode. Episodes are counted as failures if the robot misses
the grasp, drops the object, places the object in the wrong region, causes an unrecoverable collision,
or reaches the time limit. Manual judgments are based on direct observation and recorded evaluation
videos. No evaluation episodes are used for training or hyperparameter selection.

\paragraph{Limitations of the real-world benchmark.}
The real-world benchmark is limited to four tabletop tasks on a single robot platform and a single
workspace setup. It is therefore used to test whether the matched-capacity trend observed in LIBERO
also appears under physical execution, rather than to establish general real-world robustness across
robots, camera placements, object categories, or environments.

\section{Additional Update-Direction Visualizations}
\label{app:update_direction_all}

\begin{figure*}[t]
    \centering

    \begin{subfigure}[t]{0.95\textwidth}
        \centering
        \includegraphics[width=\linewidth]{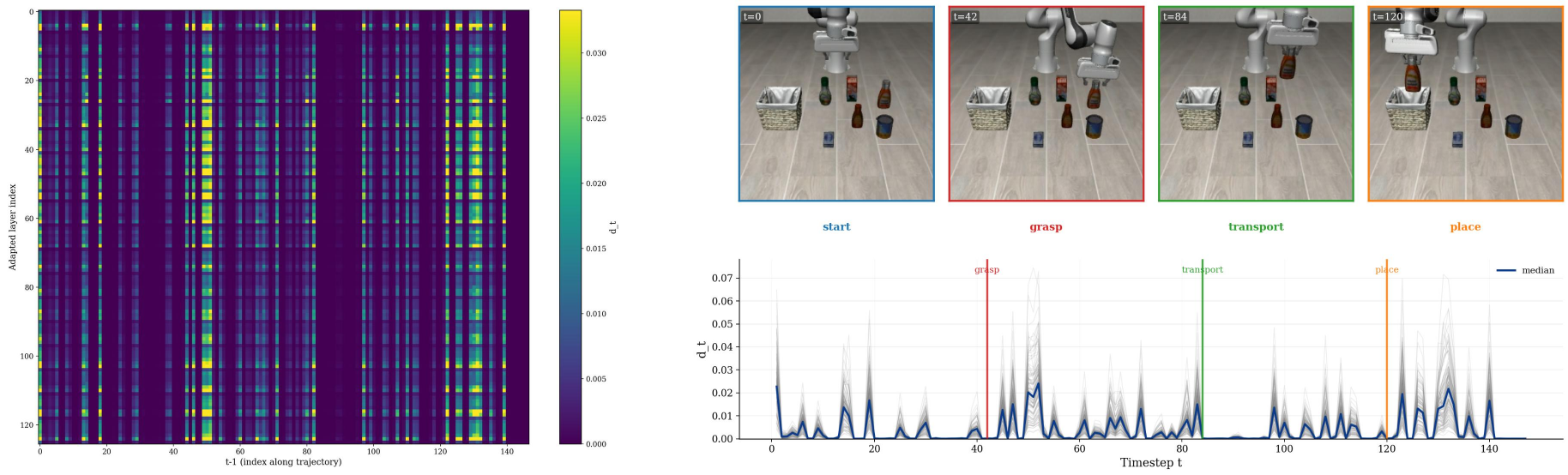}
        \caption{LIBERO-Object}
        \label{fig:update_direction_object_app}
    \end{subfigure}

    \vspace{0.6em}

    \begin{subfigure}[t]{0.95\textwidth}
        \centering
        \includegraphics[width=\linewidth]{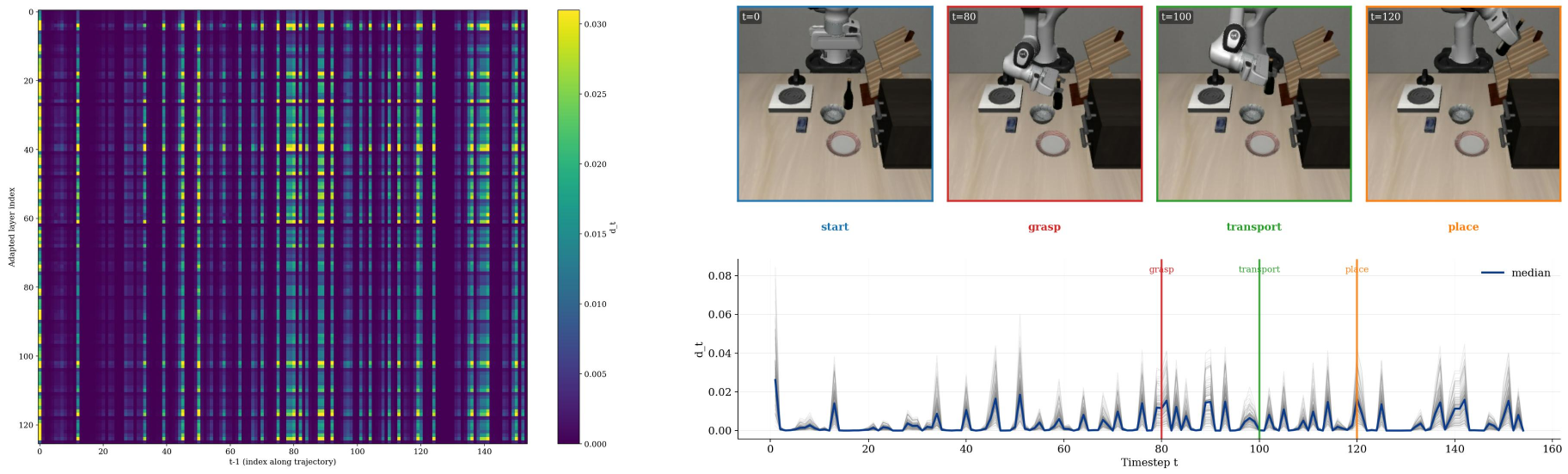}
        \caption{LIBERO-Goal}
        \label{fig:update_direction_goal_app}
    \end{subfigure}

    \vspace{0.6em}

    \begin{subfigure}[t]{0.95\textwidth}
        \centering
        \includegraphics[width=\linewidth]{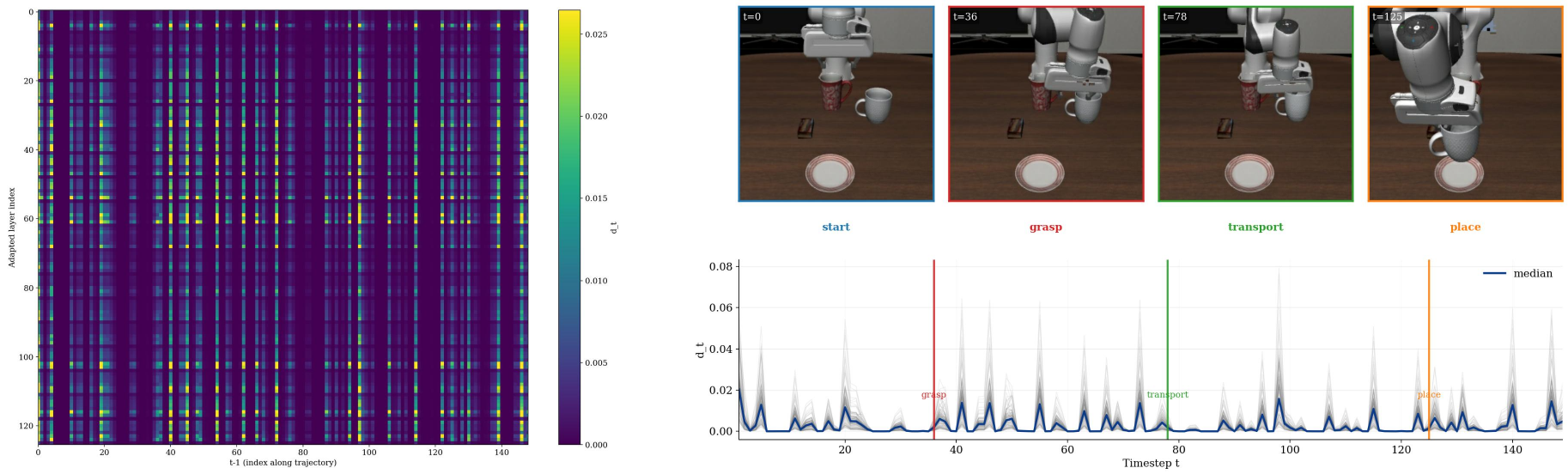}
        \caption{LIBERO-10}
        \label{fig:update_direction_10_app}
    \end{subfigure}

    \caption{\textbf{Additional update-direction visualizations for the remaining LIBERO suites.} Each row shows one representative rollout from LIBERO-Object, LIBERO-Goal, or LIBERO-10. In each row, the left panel shows the adjacent-step update-direction distance $d_t$ across adapted layers and timesteps, the upper-right row shows four rollout frames, and the lower-right panel overlays the per-layer distances with their across-layer median. Vertical lines indicate the displayed grasp, transport, and placement events. As in the main-text LIBERO-Spatial example, large $d_t$ peaks are sparse and tend to occur near these transitions.}
    \label{fig:update_direction_appendix}
\end{figure*}

Figure~\ref{fig:update_direction_appendix} provides additional update-direction visualizations
for LIBERO-Object, LIBERO-Goal, and LIBERO-10. Consistent with the LIBERO-Spatial example
in the main text, update-direction changes are sparse over time, coherent across multiple adapted
layers, and tend to occur near salient manipulation events such as grasping, transport transitions,
and placement.

\section{Broader impacts}
PhaseLoRA aims to make downstream adaptation of continuous-action VLA policies more parameter-efficient and effective. Potential positive impacts include reducing the computational cost of adapting robot policies and making manipulation systems more practical for research and applications such as assistive robotics, laboratory automation, and manufacturing. At the same time, improving robot policy adaptation may lower the barrier to deploying robots in physical environments, where failures can cause material damage or safety risks if systems are used outside their validated operating conditions. The method should therefore be evaluated with task-specific safety checks, human oversight, and deployment constraints before use in real-world settings. We do not use personally identifiable data or human-subject data in the reported experiments, but broader deployment of robotic systems may raise labor, safety, and accountability concerns that depend on the application context.

\section{Existing assets and licenses}
\label{app:licenses}
We use several existing assets and follow their corresponding licenses and terms of use.
The LIBERO benchmark is used for the main controlled experiments; we cite the original
LIBERO paper and use the official LIBERO datasets and task annotations. According to
the official LIBERO resources, the LIBERO codebase is released under the MIT License
and the datasets are released under Creative Commons Attribution 4.0 International
(CC BY 4.0). We use the standard LIBERO-Spatial, LIBERO-Object, LIBERO-Goal, and
LIBERO-10 suites.

Our experiments build on the public OpenPI implementation and the public \(\pi0.5\)
base checkpoint. We cite the original \(\pi0.5\) work and initialize from the public base
checkpoint described in Appendix~\ref{app:backbone}. OpenPI is released under the Apache License 2.0,
and the repository also includes Gemma license and terms files for Gemma-based
components. We do not redistribute the original pretrained checkpoint or LIBERO
datasets in our supplementary code release. Instead, the supplementary material provides
instructions for obtaining these assets from their original sources and for using them in
accordance with their respective licenses and terms of use.

%%%%%%%%%%%%%%%%%%%%%%%%%%%%%%%%%%%%%%%%%%%%%%%%%%%%%%%%%%%%

%\newpage
%\input{checklist.tex}

\end{document}